%% file: main.tex
\documentclass[10pt,twocolumn,letterpaper]{article}
\usepackage[pagenumbers]{cvpr} 
\input{preamble}
\definecolor{cvprblue}{rgb}{0.21,0.49,0.74}
\usepackage[pagebackref,breaklinks,colorlinks,allcolors=cvprblue]{hyperref}
\usepackage{makecell}

\def\paperID{6318}
\def\confName{CVPR}
\def\confYear{2025}

\title{Live\raisebox{-0.25\height}{\includegraphics[scale=0.5]{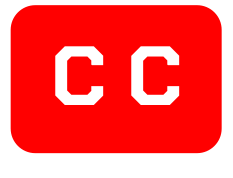}}: Learning Video LLM with Streaming Speech Transcription at Scale}
\author{Joya Chen$^{1*}$ \authorskip Ziyun Zeng$^{1*}$ \authorskip Yiqi Lin$^{1*}$ \authorskip Wei Li$^2$ \authorskip Zejun Ma$^2$ \authorskip Mike Zheng Shou$^{1}$\textsuperscript{\Letter}\\[1mm]
$^1$Show Lab, National University of Singapore \quad $^2$ByteDance
}
\begin{document}
\twocolumn[{
\maketitle
\vspace{-3em}
\renewcommand\twocolumn[1][]{#1}
\begin{center} 
    \centering
    \includegraphics[width=1.0\textwidth]{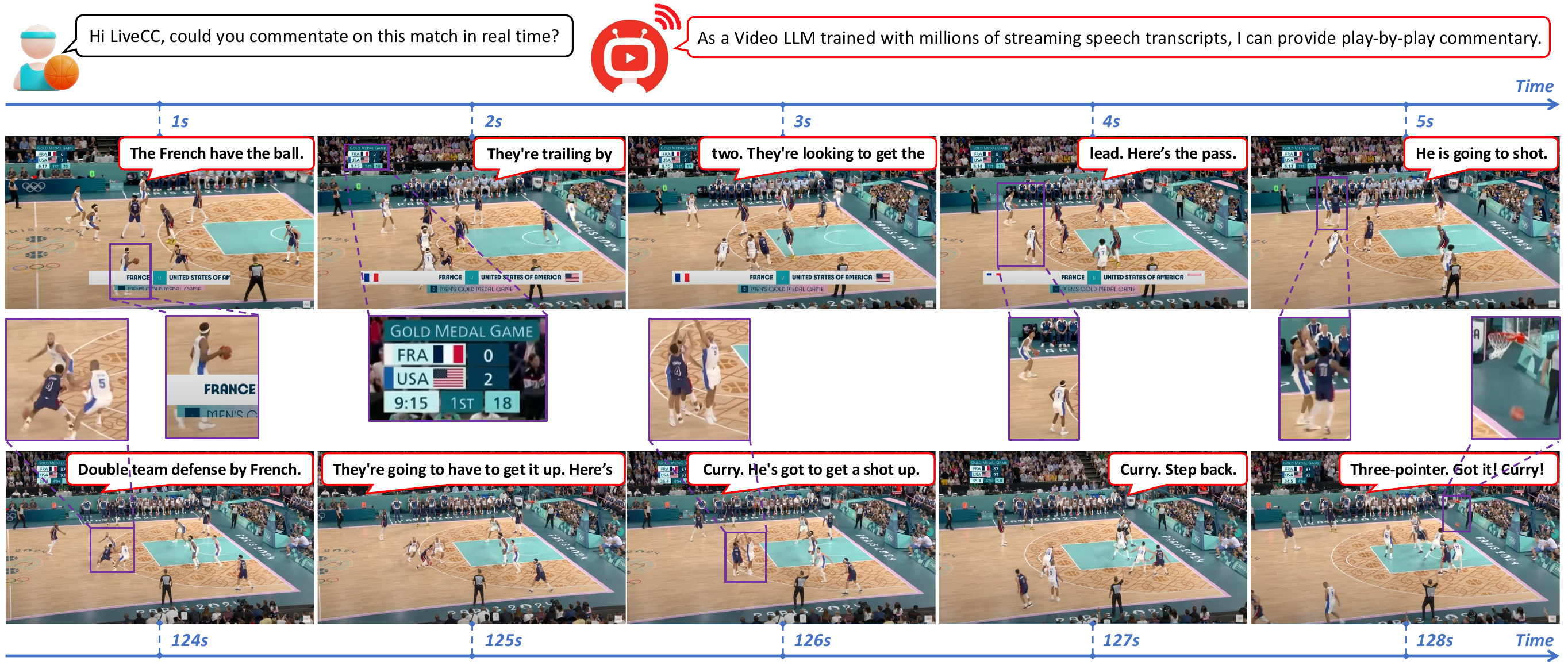}
    \vspace{-20pt}
    \captionof{figure}{LiveCC provides real-time commentary for streaming video, emulating a human  commentator. This example is drawn from the YouTube video (ID: \href{https://www.youtube.com/watch?v=I7pTpMjqNRM}{\texttt{I7pTpMjqNRM}}), featuring the Paris 2024 Olympics Men's Basketball Final between France and the USA. Our 7B model generates continuous commentary with a latency of less than 0.5 seconds per frame, supporting real-time applications at 2 FPS.}
    \label{figure:teaser}
\end{center}
}]

\footnotenosuper{\textsuperscript{\Letter}Corresponding Author. \quad \textsuperscript{*}Equal Contribution.}

\maketitle
\input{sec/0_abstract}    
\input{sec/1_intro}

\input{sec/2_related}

\input{sec/3_method}

\input{sec/4_experiments}
\input{sec/5_conclusion}

\section*{Acknowledgments}

This research is supported by the National Research Foundation, Singapore, under its AI Singapore Programme (AISG Award No: AISG3-RP-2022-030). Joya Chen proposed the idea, designed and implemented the data production pipeline and the model training/inference codebase. Ziyun Zeng built the benchmark and designed and implemented the free-form commentary evaluation. Yiqi Lin made significant contributions to baseline model analysis, data collection, and paper writing. We jointly trained and evaluated the models and discussed the results together. We acknowledge the resource support from TikTok AIIC and thank Wei Li and Zejun Ma for their valuable guidance. We are also grateful for the insightful discussions with Rui Qian and Yu Li, and we thank Kevin Qinghong Lin, Zhaoyang Lv, Yichi Zhang, and Huiyu Wang for their constructive comments.

{
    \small
    \bibliographystyle{ieeenat_fullname}
    \bibliography{ai}
}

\input{sec/X_suppl}

\end{document}

%% file: preamble.tex
\usepackage{colortbl}
\usepackage{xcolor}
\usepackage{booktabs}  
\usepackage{graphicx, amsmath, amssymb, multirow, overpic, textpos}
\usepackage{array, color}
\usepackage{amsmath}
\usepackage{bbm}
\usepackage{algorithm}
\usepackage{algorithmic}

\usepackage{pifont}

\definecolor{citecolor}{HTML}{0071bc}
\definecolor{color_ao}{gray}{0.5}
\definecolor{color_our}{rgb}{0.66,0.82,0.56}
\definecolor{color_pre}{rgb}{0.52,0.59,0.69}
\definecolor{Gray}{gray}{0.9}
\definecolor{LighterGray}{gray}{0.93}
\definecolor{LightGrayForTableRule}{gray}{0.92}
\definecolor{DarkGray}{gray}{0.5}
\definecolor{Black}{rgb}{0.0, 0.0, 0.0}
\definecolor{NiceBlue}{rgb}{0.11764705882352941, 0.5647058823529412, 1.0}
\definecolor{NiceGreen}{rgb}{0.0, 0.5, 0.0}

\definecolor{citecolor}{HTML}{0071BC}
\definecolor{linkcolor}{HTML}{ED1C24}
\usepackage{tabularx}
\newcommand{\tablestyle}[2]{\setlength{\tabcolsep}{#1}\renewcommand{\arraystretch}{#2}\centering\footnotesize}

\definecolor{LightGrayForTableRule}{gray}{0.92}

\newcommand{\authorskip}{\hspace{5mm}}
\usepackage[hang,flushmargin]{footmisc} 

\makeatletter
\newcommand{\footnotenosuper}[1]{%
   \def\@makefnmark{\hbox{}}
   \footnote{#1}
   \def\@makefnmark{\hbox{\@textsuperscript{\normalfont\@thefnmark}}}
}
\makeatother

\usepackage{marvosym}

\usepackage{color}
\usepackage{cinzel}

\definecolor{verylightblue}{RGB}{220,235,245}
\newcommand{\baseline}[1]{\cellcolor{verylightblue}{#1}}
\definecolor{lightgray}{RGB}{211,211,211}

\usepackage[most]{tcolorbox} 
\usepackage{xcolor} 
\usepackage{lipsum}

\newtcolorbox{UserBox}[2][]{
    colback=green!5!white,
    colframe=green!55!black,
    coltext=black,
    boxsep=1pt,
    arc=5pt,
    auto outer arc,
    left=5pt,
    right=5pt,
    top=1pt,
    bottom=1pt,
    box align=top,
    title=#2,
    #1
}
\newtcolorbox{AssistantBox}[2][]{
    colback=orange!5!white,
    colframe=orange!75!black,
    coltext=blue!25!black,
    boxsep=1pt,
    arc=5pt,
    auto outer arc,
    left=5pt,
    right=5pt,
    top=1pt,
    bottom=1pt,
    box align=top,
    title=#2,
    #1
}
\newtcolorbox{VideoInputBox}{
    colback=gray!5!white,
    colframe=gray!75!black,
    coltext=gray!25!black,
    boxsep=5pt,
    arc=5pt,
    left=5pt,
    right=5pt,
    top=0.5pt,
    bottom=0.5pt,
    box align=top,
    before upper=\hspace{-1em}\textbf{Streaming video input: }
}

\usepackage{xcolor} % For color customization
\usepackage{listings}

%% file: sec/0_abstract.tex
\definecolor{datasetcolor}{HTML}{9467BD}    % Deep purple
\definecolor{modelcolor}{HTML}{D62728}      % Red for models
\definecolor{benchmarkcolor}{HTML}{2CA02C}  % Green for benchmarks

% Hyperlink commands
\newcommand{\datasetlink}[2]{\href{#1}{\textcolor{datasetcolor}{\texttt{#2}}}}
\newcommand{\modellink}[2]{\href{#1}{\textcolor{modelcolor}{\texttt{#2}}}}
\newcommand{\benchmarklink}[2]{\href{#1}{\textcolor{benchmarkcolor}{\texttt{#2}}}}

\begin{abstract}
Recent video large language models (Video LLMs) often depend on costly human annotations or proprietary model APIs (\emph{e.g.}, GPT-4o) to produce training data, which limits their training at scale. In this paper, we explore large-scale training for Video LLM with cheap automatic speech recognition (ASR) transcripts. Specifically, we propose a novel streaming training approach that densely interleaves the ASR words and video frames according to their timestamps. Compared to previous studies in vision-language representation with ASR, our method naturally fits the streaming characteristics of ASR, thus enabling the model to learn temporally-aligned, fine-grained vision-language modeling. To support the training algorithm, we introduce a data production pipeline to process YouTube videos and their closed captions (CC, same as ASR), resulting in \datasetlink{https://huggingface.co/datasets/chenjoya/Live-CC-5M}{\texttt{\underline{Live-CC-5M}}} dataset for pre-training and \datasetlink{https://huggingface.co/datasets/chenjoya/Live-WhisperX-526K}{\texttt{\underline{Live-WhisperX-526K}}} dataset for high-quality supervised fine-tuning (SFT). Remarkably, even without SFT, the ASR-only pre-trained  \modellink{https://huggingface.co/chenjoya/LiveCC-7B-Base}{\texttt{\underline{LiveCC-7B-Base}}} model demonstrates competitive general video QA performance and exhibits a new capability in real-time video commentary. To evaluate this, we carefully design a new \benchmarklink{https://huggingface.co/datasets/stdKonjac/LiveSports-3K}{\texttt{\underline{LiveSports-3K}}} benchmark\textcolor{citecolor}{\textsuperscript{\textnormal{1}}}, using LLM-as-a-judge to measure the free-form commentary. Experiments show our final \modellink{https://huggingface.co/chenjoya/LiveCC-7B-Instruct}{\texttt{\underline{LiveCC-7B-Instruct}}} model can surpass advanced 72B models (Qwen2.5-VL-72B-Instruct, LLaVA-Video-72B) in commentary quality even working in a real-time mode. Meanwhile, it achieves state-of-the-art results at the 7B/8B scale on popular video QA benchmarks such as VideoMME and OVOBench, demonstrating the broad generalizability of our approach. All resources of this paper have been released at \href{https://showlab.github.io/livecc}{showlab.github.io/livecc}.
\footnotenosuper{\textcolor{citecolor}{\textsuperscript{\textnormal{1}}}Welcome to participate the real-time video commentary competition at CVPR25 workshop \href{https://sites.google.com/view/loveucvpr25/track2}{loveucvpr25/track2}. Free GPT-4o judging provided.}
\end{abstract}

%% file: sec/1_intro.tex
\section{Introduction}
\label{sec:intro}
The success of large language models (LLMs)~\cite{chatgpt,gpt4,llama1,llama2,llama3,gemini,qwen,qwen2} owes much to the large-scale auto-regressive language pre-training~\cite{scaling_law,chinchilla,gpt1,gpt2,gpt3}. This inspires large multimodal models (LMMs)~\cite{llava,minigpt4,gpt4v,gpt4o,qwen_vl}, which are initially only achieved by small-scale instruction tuning~\cite{llava,minigpt4}, to increasingly emphasize data scaling during their training. While early large multimodal models (LMMs) such as LLaVA~\cite{llava} were only supervised fine-tuned on 158K image QA samples, recent advanced approaches~\cite{onevision,vila,mm1,llava-video-178k,internvl2.5} have expanded the training data to millions of multimodal conversation samples, substantially benefiting from the increased data size.

A long-term ambition in this field is to develop LMMs akin to \emph{J.A.R.V.I.S.}, seamlessly assisting humans in real-life scenarios. Building on prior successes in LLMs/LMMs, a possible way is to collect extensive streaming video-text chat data. For instance, recordings of a basketball coach providing real-time feedback to a novice player could be great data for training. However, previous studies on streaming video LLMs~\cite{live,flashvstream,streaming_dvc,dispider,streamchat+bench,videollama3,streamchat,mmduet,videollamb,vita,vita1.5,videostreaming} have explored the difficulties of collecting and scaling up such data. They either rely on LLMs to generate ``hallucinated" streaming conversations from video annotations, or fine-tune on small-scale dense caption datasets~\cite{activitynet_captions,youcook2,vitt}. Neither approach is scalable enough to yield a truly powerful streaming video LLM.

To address these limitations, two primary approaches merit consideration. First, recent video-text datasets~\cite{llava-video-178k,panda70m,sharegpt4video,internvid} increasingly employ advanced LMMs, such as GPT-4o~\cite{gpt4o}, for synthetic data generation. While effective, this approach is costly and risks violating usage terms. Another alternative leverages the inherent audio channel in videos by utilizing automatic speech recognition (ASR) transcriptions as textual data. Prior works~\cite{howto100m,yttemporal,hdvila,vid2seq,vidchapters} have explored large-scale video-ASR  learning but typically treat ASR transcriptions as global video captions, overlooking their valuable temporal alignment. In practice, some ASR texts naturally synchronize with visual content, offering an untapped opportunity for video-language learning, escepically for streaming applications.

In this work, we aim to scale video LLM training by ASR transcriptions. We propose a novel streaming training approach that densely interleaves ASR words with corresponding video frames, as illustrated in Figure~\ref{fig:method}. The model is trained to generate frame assigned ASR words in an autoregressive manner. This approach marks a significant departure from prior LMMs~\cite{llava,minigpt4,live,streaming_dvc,qwen_vl}, which primarily learn from complete sentences or paragraphs. In contrast, our method simply learns the native \textit{short}, \textit{incomplete} ASR word sequences that are temporally aligned with video frames. This offers three key advantages: 1) it aligns naturally with the real-world data, making it readily applicable to video platforms like YouTube; 2) it enables the model to learn fine-grained temporal correlations between visual content and spoken language; and 3) during inference, it facilitates seamless streaming by generating only a few words per frame, ensuring extremely low latency.

To achieve this goal, we address three fundamental challenges: 1) How can video-ASR data be effectively curated and selected for training? 2) How should video-ASR streaming sequences be efficiently modeled? 3) How can streaming word generation—termed real-time video commentary—be rigorously evaluated? To tackle these challenges, we first design a data collection pipeline that integrates cost-effective techniques to enhance ASR quality and improve visual-text alignment, such as active speaker detection~\cite{lightasd} for filtering low-quality talking-head videos. This pipeline enables the construction of the Live-CC-5M pre-training set and the Live-WhisperX-526K SFT set. Next, we incorporate our streaming pre-training approach into the Qwen2-VL-7B-Base~\cite{qwen2vl} base model, yielding LiveCC-7B-Base, and investigate key factors influencing accurate ASR word prediction, such as leveraging video title and previous ASR as context to mitigate the learning ambiguity. Then, we introduce LiveSports-3K, a new benchmark that employs the LLM-as-a-judge~\cite{chatbot_arena} framework for evaluating real-time video commentary.
We fine-tune LiveCC-7B on Live-WhisperX-526K and LLaVA-Video-178K~\cite{llava-video-178k} to obtain LiveCC-7B-Instruct, achieving state-of-the-art performance on general QA and streaming commentary tasks.

Extensive experiments demonstrate that our streaming pre-training approach on Live-CC-5M substantially enhances commentary quality and yields  improvements in general video QA performance. By fine-tuning our pre-trained model using the Live-WhisperX-526K dataset in conjunction with LLaVA-Video-178K, our method achieves state-of-the-art results on popular video QA benchmarks such as Video-MME~\cite{videomme} and OVOBench~\cite{ovobench}, as well as our proposed LiveSports-3K benchmark, and delivers competitive performance on MVBench~\cite{mvbench}. These results indicate that our comprehensive framework is not only for real-time video commentary but also beneficial to common video understanding capability.

%% file: sec/2_related.tex
\section{Related Work}
\label{sec:related_work}

\noindent\textbf{Large Multimodal Models.} 
Early LMMs~\cite{flamingo,blip2,llava,minigpt4,instructblip} achieve image dialogue by projecting the visual embedding (\eg, from CLIP~\cite{clip,siglip}) to align with LLM token embedding space. Then, lots of efforts explore more free-form interleaved vision-text chatting~\cite{gpt4v, qwen_vl, otter, gemini,deepspeed_visualchat,llava1.5,onevision,vila}, spatial/temporal grounding~\cite{gpt4roi,ferret,minigpt4v2, qwen2vl,lisa,univtg,timechat}, video comprehension~\cite{videochatgpt,videochat,vid2seq,videollama,moviechat,videollava,llavavideo,vita,gpt4o,gemini1.5,longvu}, etc. Our model is also an LMM, but it offers new insights into cost-effective and scalable ASR training data, as well as a new capability of real-time video commentary.

\noindent\textbf{Training Video LLMs.}
Popular video LLMs~\cite{videollama3,qwen2vl,qwen2.5vl,llavavideo,internvl2.5} typically rely on human- or LLM-crafted video caption/QA sequences for training. In contrast, our work focuses on training less explored ASR transcription data, leveraging its scalability and automatic extraction capabilities. Several studies~\cite{yttemporal,howto100m,hdvila,vid2seq,vidchapters, egovlp} have investigated learning spatio-temporal representations through video-ASR pre-training. The most related work is Vid2Seq~\cite{vid2seq}, which pre-trains a model to predict timestamped ASR paragraph for videos. However, its training paradigm still aligns with previous video captioning, aiming to predict an overall event. In contrast, our approach aligns with the streaming ASR, learning to predict short, incomplete ASR words per frame causally, thereby enabling more fine-grained spatial-temporal learning.

\begin{figure}[t]
    \centering
    \includegraphics[width=1.0\linewidth]{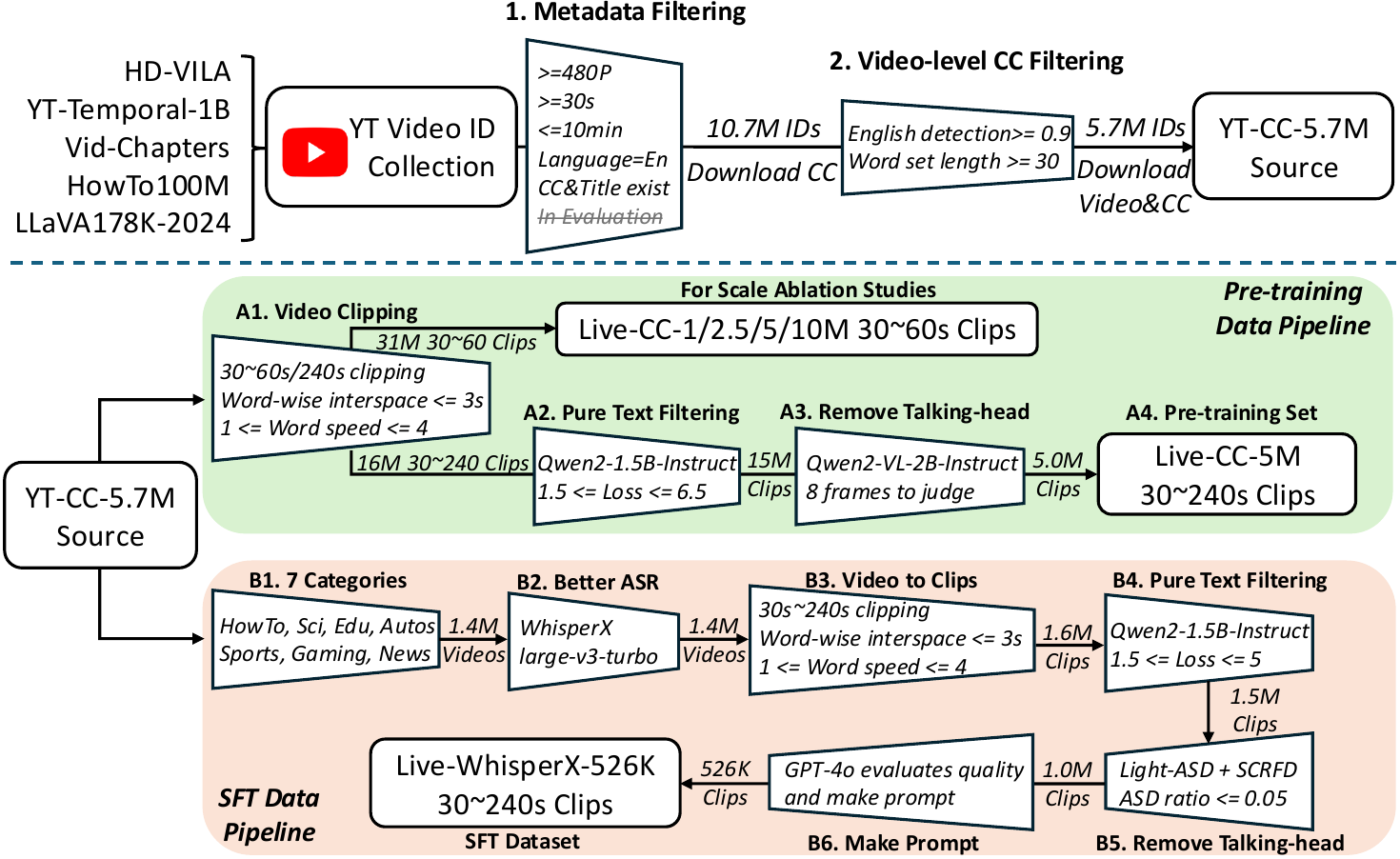}
    \vspace{-15pt}
    \caption{LiveCC data production pipeline. We begin by integrating several large-scale YouTube video datasets~\cite{hdvila,yttemporal,vidchapters,howto100m,llava-video-178k}, followed by metadata filtering, resulting in a curated pool of 5.7M videos. Then, the pre-training dataset is built using the original YouTube CC, while the SFT dataset leverages higher-quality ASR transcriptions generated by WhisperX~\cite{whisperx,whisper}. We also introduce a set of efficient filtering techniques to improve the SFT data quality. Please refer to Section~\ref{sec:data} for details.} 
    \label{fig:data_pipeline}
\end{figure}

\noindent\textbf{Streaming Video Understanding.} 
Traditional video understanding benchmarks~\cite{i3d,kinetics600,sthv1,activitynet,thumos14,tvqa,egoschema,nextqa,viddial} allow models to access entire video frames before making predictions, a setting commonly referred to as ``offline''. However, this paradigm does not align well with many real-time applications (\eg, AR glasses). Previous online video understanding tasks, such as online action detection~\cite{oad,zhao2022real,zhao2023streaming}, localization~\cite{buch2017sst,singh2017online,kang2021cag}, and captioning~\cite{streaming_dvc}, primarily focus on densely identifying current or future actions. Recent advancements in streaming video LLMs~\cite{live,streaming_dvc,dispider,videollama3,streamchat,mmduet,videollamb,vita,vita1.5,videostreaming} and benchmarks~\cite{streamchat+bench,streamingbench,ovbench,ovobench} have introduced capabilities such as proactive response, long-form streaming, and interactive multimodal conversation. However, they heavily rely on manual or GPT crafted data, and will be ``blind'' for video input before the text/audio generation finished. Our work provides a comprehensive solution that leverages ASR data to enhance both general QA and streaming capabilities, and achieves novel real-time video commentary feature, along with a new benchmark LiveSports-3K-CC for evaluation.

%% file: sec/3_method.tex
\section{Methodology}
\label{sec:method}

\subsection{Video-ASR Data Curation}
\label{sec:data}

To demonstrate the scalability of our pre-training strategy, we aggregate four recent large-scale video datasets--HD-VILA~\cite{hdvila}, YT-Temporal-1B~\cite{yttemporal}, VidChapters~\cite{vidchapters}, and HowTo100M~\cite{howto100m}--as our video sources. As illustrated in Figure~\ref{fig:data_pipeline}, we begin by retrieving video metadata (e.g., title, duration, category) and the corresponding YouTube closed captions (CC) using the released video IDs. To ensure high visual quality, we retain only videos with a resolution of at least 480p. For storage efficiency, we filter videos to be between 30 seconds and 10 minutes in length. We further require the presence of both CC and title metadata. Additionally, we observe that English videos typically yield better ASR quality; therefore, we restrict our selection to English-language content. Applying these filtering criteria results in a curated set of 10.7 million YouTube video IDs.

\begin{figure}[t]
    \centering
    \includegraphics[width=1.0\linewidth]{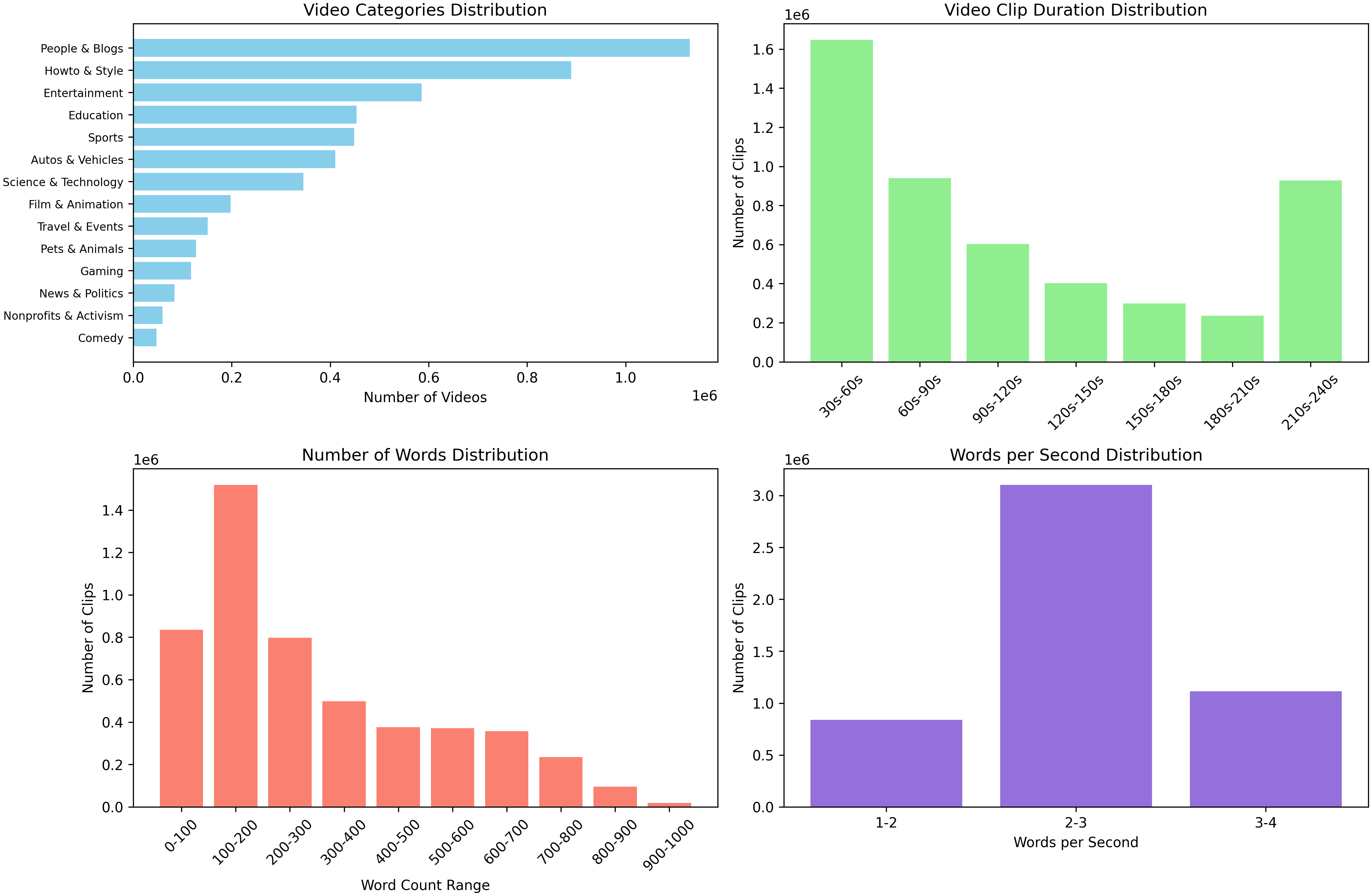}
    \vspace{-20pt}
    \caption{Overview of our proposed Live-CC-5M dataset.}
    \label{fig:pretrain_stat}
\end{figure}

\noindent\textbf{YT-CC-Source-5.7M.}
We further observed that YouTube metadata often mislabels the language category, \eg, marking videos as English despite containing code-mixed content or garbled characters. To address this, we apply the XLM-RoBERTa~\cite{xlm_roberta} (\textit{papluca/xlm-roberta-base-language-detection}) for English detection, using a confidence threshold of 0.9. In addition, we discard video IDs with sparse CC, \eg music videos with only a few words. We require each video to contain at least 30 distinct words in its CC. Applying these filters, we download these 5.7 million videos with English CC, which serves as the source for both pre-training and SFT datasets. 

\noindent\textbf{YT-CC-Source-5.7M$\rightarrow$Live-CC-5M.} 
Upon inspection, we observed that YouTube CC is generally of low quality-- lacking punctuation, case-insensitive, and frequently containing garbled characters. Nevertheless, due to their accessibility and low cost, they offer a scalable data source, making them more suitable for pre-training rather than SFT.

Therefore, we design the following steps for data curation:
\textbf{A1)} 
First, we segment the video based on ASR word timestamp gaps. If the gap between words exceeds 3 seconds, a new clip is generated. If a clip exceeds the maximum length, it is split into a new clip. 
Clips shorter than 30 seconds or with a speech rate outside the 1 to 4 words per second range are discarded.
The extended range of silence or abnormal speech speed makes it hard for the model to learn the end-of-sequence (EOS) predictions.
For pre-training, the default maximum clip length is set to 240 seconds.
For ablation studies, the clip length is set to 60 seconds for training efficiency. We rank these clips by their word set size, which reflects content informativeness, and create pre-training subsets with 1M, 2.5M, 5M, and 10M clips. 
\textbf{A2)} We compute the pure text loss of ASR transcripts by language model to assess their dependency on visual content. A very low perplexity suggests the transcript is self-contained and does not require visual grounding, while a very high perplexity often correlates with poor ASR quality. Empirically, we use Qwen2-1.5B-Instruct~\cite{qwen2} to retain samples with loss values in the range of 1.5 to 6.5;
\textbf{A3)} To remove videos with people consistently facing the camera and talking without meaningful visual information, we apply visual filtering using Qwen-VL-2B-Instruct~\cite{qwen2vl}. For each video, we use 8 uniformly sampled frames to detect persistent face-speaking content by prompting Qwen-VL-2B-Instruct.
We keep the videos if the model's confidence in detecting the talking head is below a threshold of 0.3. 

Finally, we obtain Live-CC-5M for pretraining. Figure~\ref{fig:pretrain_stat} shows the statistics of these data samples.

\begin{figure}[t]
    \centering
    \begin{subfigure}[b]{\linewidth}
        \centering
        \includegraphics[width=\linewidth]{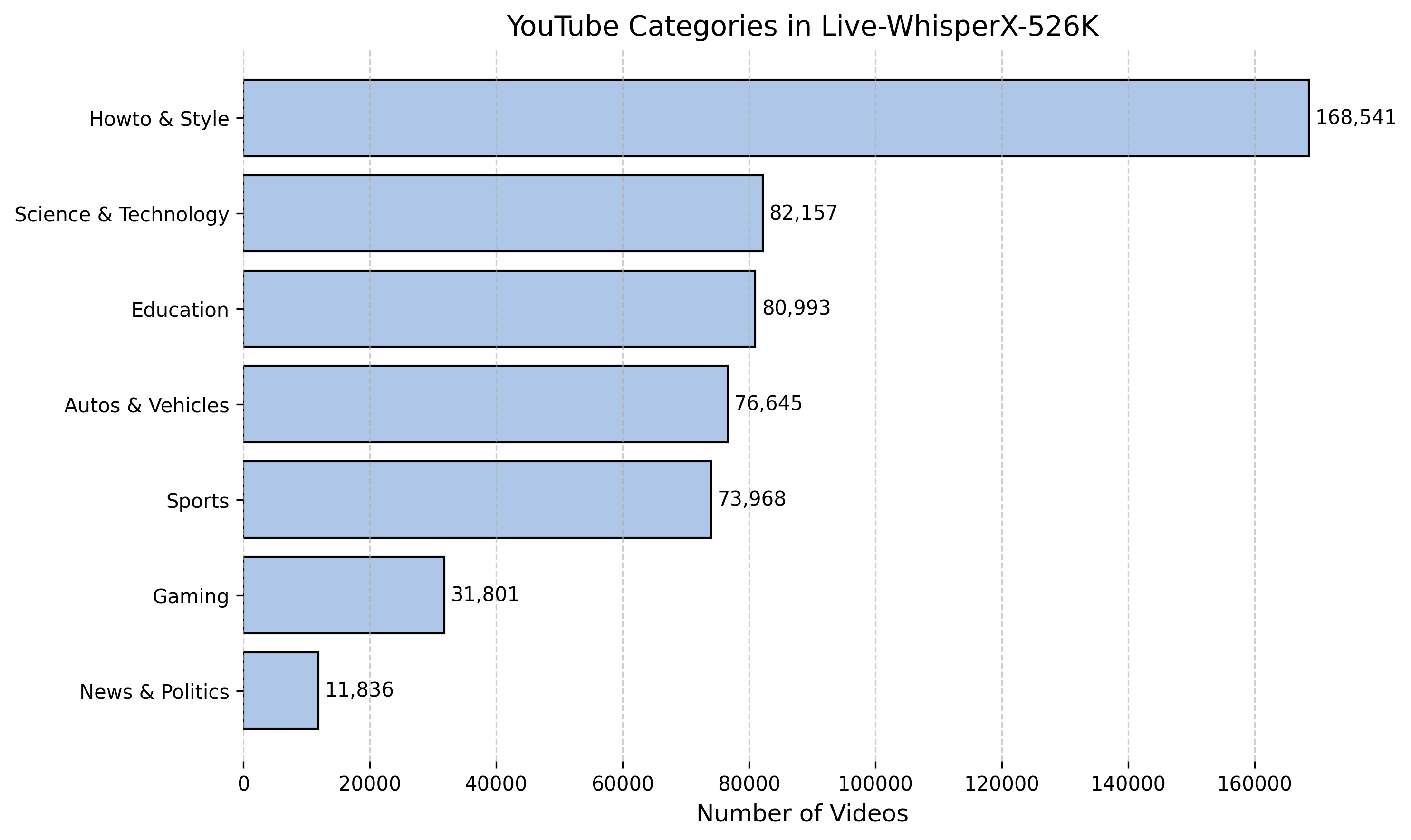}
        \vspace{-10pt}
        \caption{Statistics of our proposed Live-WhisperX-526K dataset.}
        \label{fig:sft_stat}
    \end{subfigure}
    \begin{subfigure}[b]{\linewidth}
        \centering
        \includegraphics[width=\linewidth]{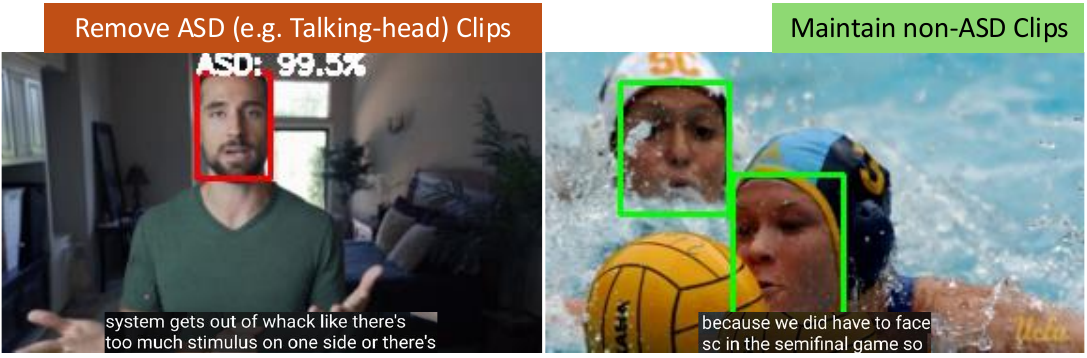}
        \vspace{-10pt}
        \caption{An example of ASD removal in SFT data pipeline.}
        \label{fig:asd_example}
    \end{subfigure}
    \begin{subfigure}[b]{\linewidth}
        \centering
        \includegraphics[width=\linewidth]{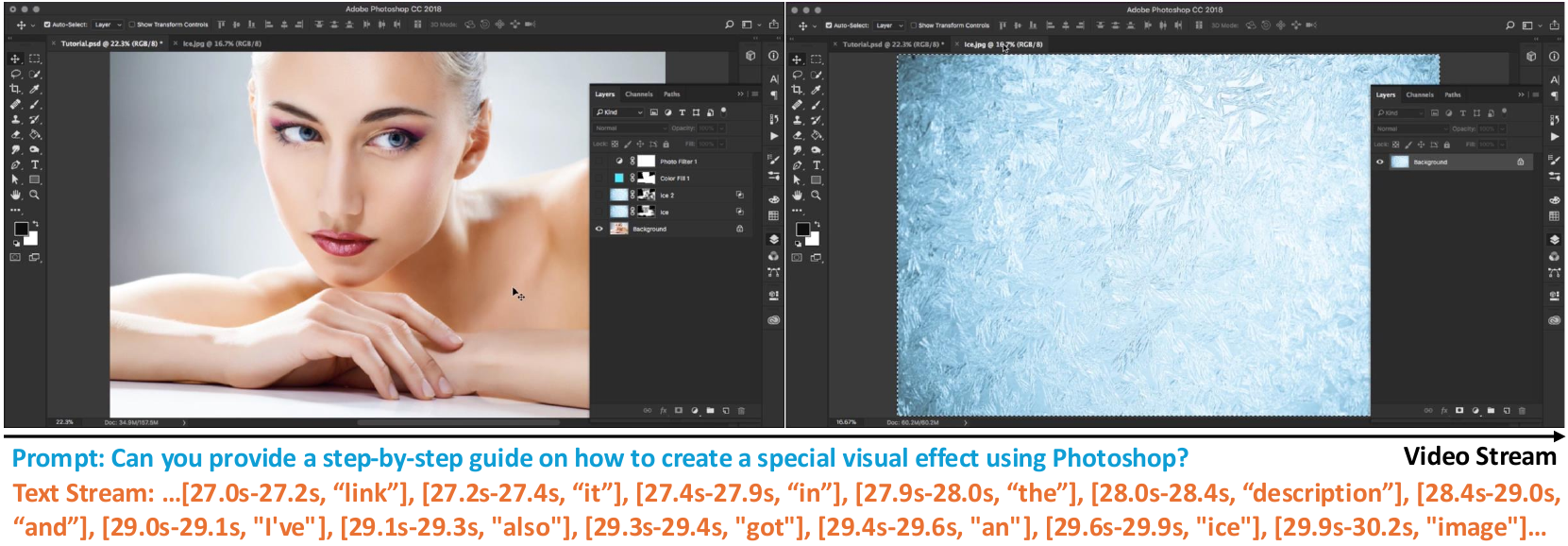}
        \vspace{-10pt}
        \caption{An example from the Live-WhisperX-526K dataset.}
        \label{fig:sft_example}
    \end{subfigure}
    \vspace{-20pt}
    \caption{Overview of the Live-WhisperX-526K dataset.}
    \label{fig:sft_data}
\end{figure}

\begin{figure*}[t]
    \centering
    \includegraphics[width=1.0\linewidth]{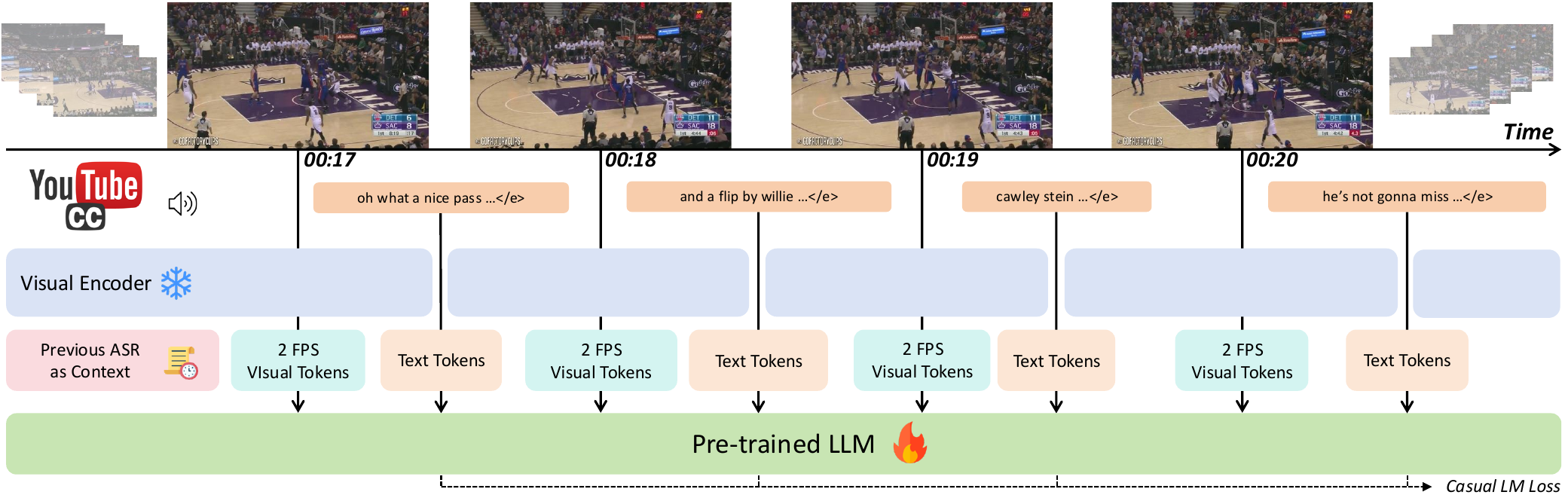}
    \vspace{-18pt}
    \caption{\textbf{Modeling Overview of LiveCC.}
     The model processes streaming video frames through a visual encoder to produce visual tokens while assigning ASR text from corresponding frame intervals as text tokens. The LLM autoregressively predicts text tokens within this densely interleaved token sequence. To mitigate learning ambiguity, additional context of preceding ASR text or video title is provided during  pre-training. During SFT, the context part is only user query to match the real-world applications.}
    \label{fig:method}
\end{figure*}
\noindent\textbf{YT-CC-Source-5.7M$\rightarrow$Live-WhisperX-526K.} 
As the low-quality YouTube CC makes them unsuitable for SFT data, we further perform the following steps to obtain high-quality, visually grounded ASR transcription:
\textbf{B1)} We only maintain 7 YouTube categories shown in Figure~\ref{fig:sft_stat} but filter out ``People \& Blogs'' and ``Film \& Animation'', as their ASR content typically lacks correspondence with the visual events;
\textbf{B2)} We employ WhisperX~\cite{whisper,whisperx} (\texttt{large-v3-turbo}) to generate more accurate, word-level aligned ASR transcriptions;
\textbf{B3)} Similar to Step A1, the maximum clip length is 240 seconds. 
For pretraining, since pre-ASR provides context, clips can be split in the middle of sentences. However, during the instruction fine-tuning stage, where no pre-ASR context is available, we ensure that each clip begins at the start of a sentence. Specifically, the last ASR word must be a period, question mark, or exclamation mark, and the current clip must start with a capital letter;
\textbf{B4)} The same as Step A2, while the range of text perplexity is 1.5 to 5;
\textbf{B5)} Despite the above filtering steps like Step A3, we observe that many remaining videos are dominated by talking-head content, which is often useless for training real-time video commentary. To address this, we employ active speaker detection (ASD)~\cite{lightasd} to identify and exclude such videos. For efficiency, we optimize Light-ASD~\cite{lightasd} pipeline in face detection, tracking, and multiprocessing, achieving a 250$\times$ speed-up. As a result, processing a 5-minute video now takes only 1–1.5 seconds. An ASD removel example is in Figure~\ref{fig:asd_example}.
\textbf{B6)} Since these ASR transcripts lack associated user prompts, we employ GPT-4o~\cite{gpt4o} to generate a prompt for each sample. The prompts are crafted to match the style and intent of the speech transcription without revealing specific content. With this prompt, we no longer need pre-ASR applied during SFT.

Finally, we get a high-quality SFT dataset comprising 526K video clips, each paired with word-level timestamped ASR transcripts and a user prompt.
Figure~\ref{fig:sft_example} shows an example in Live-WhisperX-526K dataset.

\subsection{Modeling}
\label{sec:model}

\noindent\textbf{Training with Dense Interleaving Sequence.}
As shown in Figure~\ref{fig:method}, our model architecture builds upon Qwen2-VL~\cite{qwen2vl}, which integrates a Vision Transformer~\cite{vit} with basic dynamic resolution support and uses Qwen2~\cite{qwen2} as the LLM backbone. We adopt the \textit{base} version of Qwen2-VL, which is pretrained extensively on image-text data but has limited exposure to video-text pairs. Following standard practice~\cite{llava}, the model is trained to autoregressively predict text tokens while treating visual tokens as non-predictive inputs, as illustrated in Figure~\ref{fig:method}. Unlike existing approaches that use either captioning~\cite{llava} or image-text interleaving~\cite{vila} style input, we propose to densely interleave ASR words with video frames along the temporal dimension.
The training sequence is formatted as,
\vspace{-2pt}
\begin{equation*}
\begin{aligned}
\label{eq:training_seq}
&\texttt{[Con]<F$_{t:t+k}$><W$_{t:t+k}$><F$_{t+k:t+2k}$><W$_{t+k:t+2k}$>} \\
&~~~~\texttt{...<F$_{t+n*k:t+(n+1)*k}$><W$_{t+n*k:t+(n+1)*k}$>},
\end{aligned}
\vspace{-2pt}
\end{equation*}
where \texttt{[Con]} denotes context information of the video (\eg, prompt, previous ASR, video title), \texttt{<F>} denotes a frame, \texttt{<W>} denotes the words, $t$ represents the time index and $k$ represents the time intervals. 
By default, we use 2 FPS frame rate and $k=1$ as the time interval.
We incorporate video titles and preceding ASR text as contextual information to enhance text coherence, since ASR text may start from the middle of a sentence, or use informal, verbal language.
A newline character concatenates the video title and previous ASR texts if the ASR texts are available.

\noindent\textbf{Sequence Pre-processing.}
For pre-training, we utilize the original YouTube ASR transcripts, which employ fixed timestamps to segment speech into chunks of approximately 2 to 3 seconds. To approximate word-level alignment, we uniformly distribute each segment's duration across its constituent words. This heuristic yields reasonably accurate word-level timestamps across the entire video. In contrast, during SFT, we leverage WhisperX, which provides precise word-level timestamps, as detailed in Section~\ref{sec:data}. To disambiguate the true end-of-sequence (EOS) from temporary pauses in streaming, we simply use the ellipsis token (`` ...'') as an special EOS indicator appended to the per-frame text tokens. For silent frames without corresponding subtitles, we directly predict this ellipsis token.

\noindent\textbf{Training Strategy.}
Our model training incorporates two stages including pre-training and SFT.
For the pre-training, we solely train the model with dense interleaving sequences.
The objective is to align frame features with the temporally synchronized ASR words, enabling the model to capture temporal correlations between frames and language.
Next, to improve the ability of LiveCC models to solve a diverse set of downstream tasks, we jointly train the model with our Live-WhisperX-526K in streaming mode, general video and image datasets~\cite{llava-video-178k} for common caption or QA.
To achieve this, we make the streaming training be compatible with the Qwen2-VL~\cite{qwen2vl} conversation template. The details can be found in the supplementary material.

\noindent\textbf{Inference.}
During inference, our LiveCC model processes input frames sequentially. To accelerate language decoding, we cache the Key-Value  (KV) pairs of previous prompts, visual frames, and generated text. For long sequences, we discard visual tokens every 240 seconds—consistent with the maximum duration in SFT training—while retaining the text tokens to prefill the model again.

\begin{figure*}
    \centering
    \includegraphics[width=0.97\linewidth]{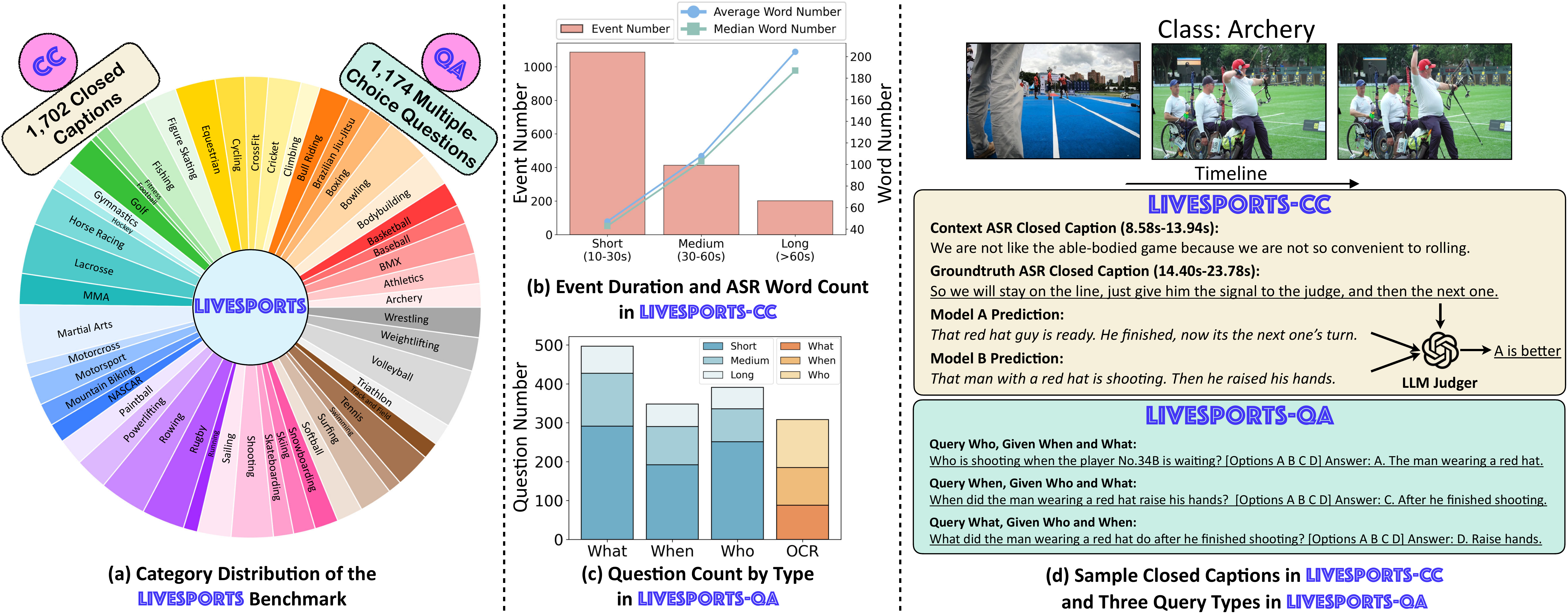}\vspace{-2pt}
    \caption{\textbf{(a)} Category Distribution of the LiveSports Benchmark: The benchmark includes 3k live CCs and MCQs, split into two tracks: CC and QA. \textbf{(b)} Event Duration and ASR Word Count in the CC track: For CC, event duration (left y-axis) and ASR word count (right y-axis) are analyzed, with durations categorized as short, medium, and long. \textbf{(c)} Question Count by Type in the QA track: Questions are grouped into three query types, with additional tracking of those requiring OCR for each type. \textbf{(d)} Sample CCs and Query Types.}
    \label{fig:benchmark}
\end{figure*}

\section{The LiveSports-3K Benchmark}
\label{sec:benchmark}

\subsection{LiveSports-3K Data Collection}
As mentioned in Section~\ref{sec:related_work}, we present \textbf{LiveSports-3K}, a comprehensive benchmark designed for systematic evaluation of video understanding models' capabilities. 
Unlike previous sports benchmarks like SoccerNet~\cite{soccernet} and MatchTime~\cite{matchtime}, which focus on specific sports, our benchmark spans a broader range of common sports to ensure generalizability.
To achieve this, we prompt GPT-4o-mini~\cite{gpt4o} to select the ongoing sports videos and classify sports categories of our Live-WhisperX-526K dataset.
We focused on the top 50 most frequent sports categories and randomly sampled 12 videos from each category, yielding a pool of 600 candidate videos covering popular sports.
After collecting candidate videos, we used GPT-4o-mini to merge ASR transcriptions into semantically coherent events, recording each event's start and end timestamps.
Given that ASR transcriptions are not always visually relevant, we recruited English-proficient annotators to filter out irrelevant events according to three criteria:
\textbf{1}) The event must last more than 10 seconds;
\textbf{2}) The event clip must contain ongoing sports action;
\textbf{3}) Most ASR transcriptions within the event clip should be visually grounded.
Events that failed to meet any of these criteria were discarded.
We curated 416 videos across 49 sports categories (one category lack of qualified videos).
The category distribution is shown in Figure~\ref{fig:benchmark}(a).
We further remove these videos from our training dataset for fair evaluation.

\subsection{Crafting LiveSports-3K-CC/QA}
\noindent \textbf{LiveSports-3K-CC.} Real-time commentary in sports videos encapsulates rich spatiotemporal semantics. For instance, the commentary on a football game often describes attackers, defenders, and their interactions in detail, making it an ideal source for streaming video understanding evaluations.
Therefore, we developed this track to assess video comprehension by evaluating the alignment between model-generated and groundtruth CCs from ASR.
Note that the data collection process has ensured that the ASR event transcriptions are visually grounded.
Thus, we directly leverage ASR transcriptions from qualified events as groundtruth. 
In summary, LiveSports-3K-CC consists of 1,702 events with high-quality live CCs.
Figure~\ref{fig:benchmark}(b) presents the distribution of events and word counts in three duration groups, showing a relatively balanced total word count (\emph{i.e.}, video count $\times$ word count per group) across groups. 
Additionally, Figure~\ref{fig:benchmark}(d) illustrates a sample event, demonstrating the highly visual-grounded nature of the ASR-transcribed CCs retained through our filtering criteria.

For evaluation, we prompt the model with the video title and the preceding CCs of the event, then record the model’s predictions. Given the challenges of directly assessing discrepancies between the predicted and groundtruth CCs, we adopt a pairwise comparison approach inspired by Chatbot Arena~\cite{chatbot_arena}. Specifically, for each pair of predictions, we use GPT-4o as a judge to select the better prediction based on the ground-truth CCs. The selection criteria include both stylistic and semantic consistency. The winning rates of different models serve as the ranking metric.

\begin{table*}[t]
\scriptsize
\centering
\begin{subtable}[b]{0.37\linewidth}
\centering
\tablestyle{2pt}{0.92} 
\resizebox{\linewidth}{!}{
\begin{tabular}{l|c|cc}
\toprule
\multirow{2}{*}{Video-ASR Sequence} & LiveSports-3K-CC & \multicolumn{2}{c}{Video-MME} \\
& \textit{Win Rate}$\uparrow$ & \textit{Overall}$\uparrow$ & \textit{Short}$\uparrow$  \\
\midrule
Caption (5M) & 14.0 & \textbf{61.1} & 69.4  \\
\baseline{Streaming} (5M) & \baseline{32.9} & \baseline{61.0} &
\baseline{\textbf{70.1}}  \\
Caption+Streaming (5M$\times$2) & \textbf{35.1} & 60.5 & 69.0 \\
\bottomrule
\end{tabular}}
\caption{\textbf{Training Paradigm during Pre-training.}}\label{tab:ab_pt_training_objective}
\end{subtable}
\hfill
\begin{subtable}[b]{0.34\linewidth}
\centering
\tablestyle{4pt}{0.92} 
\resizebox{\linewidth}{!}{
\begin{tabular}{l|c|cc}
\toprule
\multirow{2}{*}{Context} & LiveSports-3K-CC &  \multicolumn{2}{c}{Video-MME}\\
&\textit{Win Rate}$\uparrow$ & \textit{Overall}$\uparrow$ & \textit{Short}$\uparrow$ \\
\midrule
None & 14.7 & 60.7 & 69.0 \\
Title & 24.8  & 59.7 & 67.9 \\
Prev. ASR & 32.0 & \textbf{61.1} & 69.7 \\
Title $\&$ Prev. ASR &  \textbf{33.8}  & 60.7 & 69.4 \\
\baseline{Title $||$ Prev. ASR} & \baseline{32.9} & \baseline{61.0} & \baseline{\textbf{70.1}} \\
\bottomrule
\end{tabular}}
\caption{\textbf{Context during Pre-training.}}\label{tab:ab_pt_context}
\end{subtable}
\hfill
\begin{subtable}[b]{0.27\linewidth}
\centering
\tablestyle{4pt}{0.92}
\resizebox{\linewidth}{!}{
\begin{tabular}{l|c|cc}
\toprule
\multirow{2}{*}{Data} & LiveSports-3K-CC &  \multicolumn{2}{c}{Video-MME}\\
&\textit{Win Rate}$\uparrow$   & \textit{Overall}$\uparrow$ & \textit{Short}$\uparrow$ \\
\midrule
1M & 29.1 & 60.6 & 68.1 \\
2.5M & 30.8 & 60.9 & 69.1  \\
\baseline{5M} & \baseline{32.9} & \baseline{\textbf{61.0}} & \baseline{\textbf{70.1}} \\
10M & \textbf{36.0} & 58.0 & 67.6 \\
\bottomrule
\end{tabular}}
\caption{\textbf{Pre-training Scalability}.}\label{tab:ab_scale}
\end{subtable}
\vspace{-1.2em}
\caption{Ablation Study in the Pre-training Stage. \colorbox{verylightblue}{Blue} highlights our default setting. Win Rate indicates the percentage of wins against GPT-4o-08-06-generated commentary, using ground-truth ASR as the reference. Models in these table are pre-trained on a maximum of 120 frames (60 seconds at 2 FPS), so we did not show results on medium- and long- length videos for Video-MME~\cite{videomme} to avoid misjudgment.
\textbf{(a)} Both caption-style and streaming-style sequence significantly improve QA performance; however, only the streaming-style sequence yields notable gains in commentary generation.
\textbf{(b)} Incorporating previous ASR enhances both commentary and QA. In contrast, simply adding video titles degrades QA, unless previous ASR is absent (\eg, during 0–60s).
\textbf{(c)} Commentary benefits from increased pre-training data, while QA performance declines beyond 5M examples—likely due to overtraining on the single-source (streaming ASR) data.
}
\end{table*}

\noindent\textbf{LiveSports-3K-QA.}
LiveSports-3K-CC offers a valuable track for assessing video understanding comprehensively. However, it still lacks a precise criterion for analyzing model behavior, especially when errors occur.
To address this, we decompose each event into three fundamental elements:
(\textbf{i}) \textbf{When}: Captures the temporal context of the event.
(\textbf{ii}) \textbf{What}: Defines the content or action taking place in the event.
(\textbf{iii}) \textbf{Who}: Identifies the participants involved in the event.
By structuring events around these elements, we enable targeted queries for each element based on the other two, allowing us to isolate specific areas of weakness that require improvement.
Figure~\ref{fig:benchmark}(c) provides detailed examples of these three question types. Additionally, we recorded whether a question required OCR capabilities, allowing for an auxiliary evaluation of model performance on text recognition tasks. This process yielded 1,236 four-option MCQs across 414 videos, excluding two videos due to the difficulty of designing appropriate questions. Finally, we manually removed 62 questions that require speech recognition, leaving the remaining 1,174 MCQs as the final benchmark. This track includes a balanced distribution of the three query types, with OCR-reliant questions evenly distributed among them, as shown in Figure~\ref{fig:benchmark}(b).

%% file: sec/4_experiments.tex
\section{Experiments}
\label{sec:experiments}

\subsection{Experiments Setup}
\noindent\textbf{Implemetation Details.}
We initialize our model with the Qwen2-VL-7B-Base checkpoint~\cite{qwen2vl}, following most of the configurations provided in its HuggingFace release, with minimal modifications to improve efficiency. Specifically, during ablation studies of pre-training, we reduce the maximum number of frames from 768 to 120 and shorten the visual context length from 128K to 16K tokens. During formal pre-training and SFT, we increase the frame limit to 480 and extend the visual context length to 24K, while slightly lowering the minimum spatial resolution from $128\times28\times28$ to $100\times28\times28$. Pre-training ablation studies are conducted on the 30$\sim$60s Live-CC-1$\sim$10M dataset. The formal pre-training is in 30$\sim$240s Live-CC-5M. The SFT stages uses our Live-Whisper-526K and LLaVA-Video-178K~\cite{llava-video-178k} datasets (without the training set of ActivityNetQA~\cite{activitynet}, Next-QA~\cite{nextqa}, and PerceptionTest~\cite{perceptiontest}). We implement the training engine using PyTorch~\cite{pytorch} and Transformers~\cite{transformers}. The batch size for pre-training and SFT is 512 on 128 GPUs, with a learning rate of 2e-5 for pre-training and 1e-5 for SFT.

\begin{table*}[htbp]
\centering
\scriptsize
\begin{subtable}[t]{1.0\linewidth}
\centering
\tablestyle{2pt}{0.92}
\resizebox{\linewidth}{!}{
\begin{tabular}{l|l|c|ccccc|c|ccc|ccc|cc|cccc|c|c|c}
\toprule
\multirow{3}{*}{\textbf{Pre-training}} & \multirow{3}{*}{\textbf{SFT}} & \multicolumn{6}{c|}{\multirow{2}{*}{\textbf{LiveSports-3K}}} & \multicolumn{16}{c}{\textbf{VideoMME}} \\
& &  &  &  &  &  &  & \multirow{2}{*}{All} & \multicolumn{3}{c|}{Duration} & \multicolumn{3}{c|}{Perception} & \multicolumn{2}{c|}{Recognition} & \multicolumn{4}{c|}{Reasoning} & \multirow{2}{*}{OCR} & \multirow{2}{*}{Count} & \multirow{2}{*}{IS} \\
& & CC & QA & OCR & Who & When & What & & S & M & L & Te & Sp & At & Ac & Ob & Te & Sp & Ac & Ob &  &  &  \\
\midrule
\multirow{2}{*}{Qwen2-VL-7B-Base}
&LV178K & 16.7 & 67.0 & 66.1 & \textbf{70.6} & \textbf{57.6} & 71.0 & 62.7 & \textbf{74.7} & 62.4 & 51.1 & 74.5 & \textbf{61.1} & 73.4 & 64.5 & 70.1 & 49.7 & \textbf{80.4} & 49.8 & 57.0 & \textbf{76.3} & \textbf{45.1} & 76.2 \\
&\baseline{LV178K+Live526K} & \baseline{\textbf{33.7}} & \baseline{\textbf{67.1}} & \baseline{\textbf{66.8}} & \baseline{69.8} & \baseline{57.0} & \baseline{\textbf{72.3}} & \baseline{\textbf{63.6}} & \baseline{74.4} & \baseline{\textbf{63.1}} & \baseline{\textbf{53.2}} & \baseline{74.5} & \baseline{57.4} & \baseline{\textbf{75.2}} & \baseline{\textbf{66.5}} & \baseline{70.1} & \baseline{49.7} & \baseline{76.8} & \baseline{\textbf{54.4}} & \baseline{\textbf{57.5}} & \baseline{72.7} & \baseline{44.4} & \baseline{\textbf{78.9}} \\
\bottomrule
\end{tabular}
}
\caption{\textbf{Ablation study in the SFT data.}}
\label{tab:model_performance}
\end{subtable}

\begin{subtable}[t]{1.0\linewidth}
\tablestyle{2pt}{0.92}
\resizebox{\linewidth}{!}{
\begin{tabular}{l|l|c|ccccc|c|ccc|ccc|cc|cccc|c|c|c}
\toprule
\multirow{3}{*}{\textbf{Pre-training}} & \multirow{3}{*}{\textbf{SFT}} & \multicolumn{6}{c|}{\multirow{2}{*}{\textbf{LiveSports-3K}}} & \multicolumn{16}{c}{\textbf{VideoMME}} \\
& &  &  &  &  &  &  & \multirow{2}{*}{All} & \multicolumn{3}{c|}{Duration} & \multicolumn{3}{c|}{Perception} & \multicolumn{2}{c|}{Recognition} & \multicolumn{4}{c|}{Reasoning} & \multirow{2}{*}{OCR} & \multirow{2}{*}{Count} & \multirow{2}{*}{IS} \\
& & CC & QA & OCR & Who & When & What & & S & M & L & Te & Sp & At & Ac & Ob & Te & Sp & Ac & Ob &  &  &  \\
\midrule
Qwen2-VL-7B-Base
& - & 16.3   & \textbf{64.0} & \textbf{64.8} & \textbf{65.2} & \textbf{57.9} & \textbf{67.3} & \textbf{63.4} & \textbf{73.2} & \textbf{63.2} & 53.9 & \textbf{72.7} & 63.0 & \textbf{76.1} & \textbf{63.9} & \textbf{67.2} & 44.6 & 78.6 & \textbf{57.5} & 61.5 & \textbf{72.7} & 39.6 & 80.2 \\
\baseline{LiveCC-7B-Base}
& \baseline{-} & \baseline{\textbf{43.2}} & \baseline{57.9} & \baseline{61.4} & \baseline{59.4} & \baseline{50.7} & \baseline{61.9} & \baseline{61.4} & \baseline{68.1} & \baseline{58.9} & \baseline{\textbf{57.3}} & \baseline{65.5} & \baseline{63.0} & \baseline{64.9} & \baseline{60.7} & \baseline{61.0} & \baseline{\textbf{50.3}} & \baseline{\textbf{80.4}} & \baseline{56.1} & \baseline{61.5} & \baseline{61.2} & \baseline{\textbf{42.9}} & \baseline{\textbf{82.4}} \\
\midrule
Qwen2-VL-7B-Base &LV178K+Live526K & 33.7 & \textbf{67.1} & \textbf{66.8} & 69.8 & 57.0 & \textbf{72.3} & 63.6 & 74.4 & 63.1 & 53.2 & 74.5 & 57.4 & 75.2 & \textbf{66.5} & \textbf{70.1} & 49.7 & 76.8 & 54.4 & 57.5 & 72.7 & 44.4 & 78.9 \\
\baseline{LiveCC-7B-Base} &\baseline{LV178K+Live526K} & \baseline{\textbf{41.5}} & \baseline{66.8} & \baseline{66.4} & \baseline{\textbf{71.4}} & \baseline{56.1} & \baseline{70.8} & \baseline{\textbf{64.1}} & \baseline{\textbf{74.8}} & \baseline{\textbf{63.9}} & \baseline{\textbf{53.7}} & \baseline{74.5} & \baseline{\textbf{64.8}} & \baseline{74.3} & \baseline{66.1} & \baseline{68.6} & \baseline{\textbf{50.3}} & \baseline{76.8} & \baseline{52.3} & \baseline{\textbf{59.5}} & \baseline{\textbf{77.0}} & \baseline{\textbf{46.3}} & \baseline{\textbf{79.9}} \\
\bottomrule
\end{tabular}
}
\caption{\textbf{Ablation study in the SFT model initialization.}}
\end{subtable}
\vspace{-1.2em}
\caption{Ablation studies in the SFT stage. LV178K denotes the datasets used in LLaVA-Video-178K~\cite{llava-video-178k}. Live526K refers to our proposed Live-WhisperX-526K. LiveSports-3K CC denotes the win rate against commentary
generated by LLaVA-Video-72B. LiveSports-3K QA is the overall accuracy includes OCR, Who, When, What questions. Te, Sp, At, Ac, Ob, IS denotes temporal, spatial, attribute, action, object, information synopsis, respectively.}\label{tab:model_performance}
\end{table*}

\noindent\textbf{Evaluation Protocols and Metrics.}
For QA benchmarks, we evaluate our models on VideoMME~\cite{videomme}, MVBench~\cite{mvbench}, OVOBench~\cite{ovobench}, and our newly introduced LiveSports-3K-QA. For all models, we calculate the logits of multiple choices to select answers, due to the instruction following capability of streaming ASR pre-trained model has been lost. We observe this does not make difference with generation-based method~\cite{lmms-eval} for SFT models, but the evaluation is much faster.

The evaluation on LiveSports-3K-CC is like a conditioned video captioning task, where the condition comprises the video title and previous ASR text. With this condition, the model's task is to complete the ASR text based on the given video clip. Since most video LLMs lack real-time streaming capabilities, we evaluate them using a general video captioning approach, processing all video clip frames at once. In contrast, our model supports real-time inference, enabling us to generate captions on a frame-by-frame basis and then concatenate them into a complete response for evaluation. Due to the challenges of directly assessing open-ended text generation, we employ a pairwise competition approach, similar to Chatbot Arena~\cite{chatbot_arena}. We use GPT-4o~\cite{gpt4o} as the fixed competition opponent. In each competition (tested model vs. GPT-4o), we also use GPT-4o~\cite{gpt4o} acts as the judge, selecting the response that best aligns both stylistically and semantically with the ground truth ASR transcriptions. The evaluation metric is the win rate, which is defined as the proportion of times the judge favors our model over the baseline. The evaluation also involves latency comparison, which we discuss in the supplementary material.

\begin{table}[t] 
\centering 
\tablestyle{2pt}{0.92} 
\resizebox{\linewidth}{!}{
\begin{tabular}{l|cc|c|cccc} 
\toprule 
\multirow{2}{*}{\textbf{Model (7B/8B)}}  
& \multicolumn{2}{c|}{\textbf{VideoMME}} 
& \textbf{MVBench}   
& \multicolumn{4}{c}{\textbf{OVOBench}}  \\ 

& \textit{w/o sub} & \textit{w sub}  
& \textit{Avg.} 
& \textit{Avg.} & \textit{RTVP} & \textit{BT} & \textit{FAR} \\
\midrule  

LongVA-7B~\cite{longva} 
& 52.6	& 54.3 
& - 
& - & - & - & - \\

InternVL2-8B~\cite{internvl2} 
& 54.0	& 56.9
& 66.4 
& 50.2 & 60.4 & 43.4 & 46.6 \\

LLaVA-OV-7B~\cite{llavaov} 
& 58.2 & 61.5 
& 56.7 
& 52.7 & \textbf{64.0} & 43.7 & 50.5 \\ 

Oryx-7B~\cite{oryx} 
& 58.3 & 62.6 
& 63.9
& - & - & - & -  \\

mPLUG-Owl3-7B~\cite{mplug-owl3}  
& 59.3 & 68.1 
& 59.5
& - & - & - & - \\

LongVU-7B~\cite{longvu}  
& 60.6 & -
& 66.9
& 46.7 & 57.6 & 35.0 & 47.5  \\

MiniCPM-v2.6~\cite{minicpm-v} 
& 60.9 & 63.6
& -
& - & - & - & - \\ 

Qwen2-VL-7B-Instruct~\cite{qwen2vl} 
& 63.3 & 69.0 
& \textbf{67.0} 
& 50.4 & 56.0 & 46.5 & 48.7  \\

LLaVA-Video-7B~\cite{llavavideo}  
& 63.3 & 69.7 
& 58.6 
& 52.9 & 63.5 & 40.4 & \textbf{54.8} \\

\baseline{LiveCC-7B-Instruct}  
& \baseline{\textbf{64.1}} & \baseline{\textbf{70.3}}
& \baseline{62.8} 
& \baseline{\textbf{59.8}} & \baseline{59.1} & \baseline{\textbf{68.9}} & \baseline{51.5}  \\

\bottomrule 
\end{tabular} 
}
\caption{Comparison of QA accuracy (\%) across VideoMME~\cite{videomme}, MVBench~\cite{mvbench}, OVOBench~\cite{ovobench}. We only show results before the CVPR 2025 submission period (Nov, 2024).} 
\label{exp:videomme_all} 
\vspace{-0.5em} 
\end{table}

\subsection{Ablation Study}

\noindent\textbf{Pre-training Paradigm.}
We first investigate the impact of different pre-training paradigms on model performance using the following baselines: (1) Caption, where all ASR text is concatenated and appended after the visual input frames; (2) Streaming, where the model is trained on sequential frame inputs sampled at 2 FPS, predicting ASR text incrementally after receiving every 2 frames; and (3) Caption+Streaming, where each training sample contributes to both captioning and streaming objectives. As shown in Table~\ref{tab:ab_pt_training_objective}, both caption and streaming pre-training achieve strong general video QA performance (exceeding 60) on the Video-MME benchmark~\cite{videomme}, outperforming many existing SFT models. Notably, the streaming-based pre-training yields significantly better results on the commentary task compared to caption-based pre-training, highlighting the effectiveness of our proposed paradigm.

\noindent\textbf{Context Input for Pre-training.}
In Table~\ref{tab:ab_pt_context}, we observe that providing contextual information, particularly the previous ASR text, significantly improves commentary generation. This improvement stems from the fact that a 60-second segment can break the continuity of ASR, making it difficult to interpret the current segment without prior context. While incorporating the video title as additional context offers benefits on commentary, it slightly degrades performance on VideoMME. We attribute this to potential information leakage, which may make training easier. To handle the cases where no previous ASR is available (\eg, clips at the beginning of a video), we adopt a hybrid strategy: ``Title $||$ Prev. ASR'', which includes the video title only when previous ASR is unavailable. This approach strikes the best balance between enhancing commentary generation and maintaining general video QA performance.

\noindent\textbf{Pre-training Scalability.}
Table~\ref{tab:ab_scale} presents the results of scaling up the pre-training data. We observe that commentary performance consistently improves with larger data size. However, QA performance begins to decline beyond the 5M scale, likely due to the use of single-source (streaming commentary) data during pre-training. Since our primary goal is to demonstrate the effectiveness of the streaming-based pre-training, we defer the use of multi-source data to the SFT stage.

\begin{table}[t]
\centering
\scriptsize
\tablestyle{1.5pt}{0.92}
\resizebox{1.\linewidth}{!}{
\begin{tabular}{c|l|cc|ccccc}
\toprule
\multirow{2}{*}{\textbf{Size}} & \multirow{2}{*}{\textbf{Model}} 
& \multicolumn{6}{c}{LiveSports-3K $\uparrow$} \\
&  & Live? & \textbf{CC} & \textit{Overall} 
& \textit{OCR} & \textit{Who}  & \textit{When} & \textit{What} \\
\midrule

\multirow{2}{*}{\textit{\textcolor{blue}{-}}}
& GPT-4o-08-06~\cite{gpt4o}  
& \ding{55} & \textcolor{orange}{\ding{96}} & \textbf{72.2} & \textbf{74.0} & \textbf{75.8}  & \textbf{63.4}  & \textbf{75.4} \\
& Gemini-1.5-Pro~\cite{gemini}  
& \ding{55} & 52.8 
& 61.8 & 61.7 & 59.9 & 51.6 & 70.7 \\
\midrule

\multirow{5}{*}{\textit{\textcolor{blue}{72B}}}
& Qwen2-VL-72B-Instruct~\cite{qwen2vl}  
& \ding{55} & 17.0
& 70.8 & 67.8 & 74.6 & 61.2 & 74.6 \\
& VideoLLaMA-2-72B~\cite{videollama2}  
& \ding{55} & 24.8
& 62.4 & 55.7 & 63.6 & 54.3 & 67.3 \\
& LLaVA-OV-72B~\cite{llavaov}  
& \ding{55} & 29.2
& 68.7 & 61.7 & 71.1 & 61.5 & 71.8 \\
& Qwen2.5-VL-72B-Instruct~\cite{qwen2.5vl}  
& \ding{55} & 30.4 & \textbf{73.7} & \textbf{70.1} & \textbf{75.7} & \textbf{69.3} & \textbf{75.3} \\
& LLaVA-Video-72B~\cite{llavavideo} 
& \ding{55} & \textbf{35.0}
& 71.1 & 65.1 & 74.1 & 64.8 & 73.3 \\
\midrule

\multirow{9}{*}{\textit{\textcolor{blue}{7B}}}
& Qwen2-VL-7B-Instruct~\cite{qwen2vl} 
& \ding{55} & 9.3
& 65.8 & 65.8 & 67.9 & 58.8 & 69.2 \\
& Qwen2.5-VL-7B-Instruct~\cite{qwen2.5vl}  
& \ding{55} & 17.3 & 67.0 & 64.8 & 70.3 & \textbf{60.6} & 69.0\\
& InternLM-XC2.5-7B~\cite{ixc25}  
& \ding{55} & 17.3
& 59.3 & 56.7 & 60.7 & 54.9 & 61.3 \\
& Qwen2.5-Omni-7B~\cite{qwen2.5omni} (\textit{Thinker})
& \ding{55} & 17.6 & 66.8 & 66.1 & 70.0 & 60.0 & 69.2 \\
& LLaVA-Video-7B~\cite{llavavideo}  
& \ding{55} & 27.1
& 66.4 & 64.1 & \textbf{72.7} & 56.4 & 68.6 \\
& LLaVA-OV-7B~\cite{llavaov}
& \ding{55} & 27.7
& 63.4 & 60.7 & 67.4 & 53.7 & 67.1 \\
& \baseline{Qwen2-VL-7B-LiveCCInstruct}  
& \baseline{\ding{51}} & \baseline{33.7} & \baseline{\textbf{67.1}}
& \baseline{\textbf{66.8}} & \baseline{69.8} & \baseline{57.0} & \baseline{\textbf{72.3}}  \\
& \baseline{LiveCC-7B-Instruct}  
& \baseline{\ding{51}} &  \baseline{41.5}
& \baseline{66.8} & \baseline{66.4} & \baseline{71.4} & \baseline{56.1} & \baseline{70.8}  \\
& \baseline{LiveCC-7B-Base}  
& \baseline{\ding{51}} & \baseline{\textbf{43.2}}
& \baseline{57.9} & \baseline{61.4} & \baseline{59.4} & \baseline{50.7} & \baseline{61.9} \\
\bottomrule
\end{tabular}
}
\caption{Win rate on the LiveSports-3K-CC track and QA accuracy on the LiveSports-3K-QA track. GPT-4o-08-06 is used as the \textcolor{orange}{\ding{96} Baseline} for the commentary win rate comparison due to its strong performance. For a fair comparison, all Qwen models~\cite{qwen2vl,qwen2.5vl,qwen2.5omni} and our models are evaluated on a maximum of 480 frames. The Qwen models did not perform as expected, as they tend to simply caption the video rather than follow the preceding ASR context to continue the video commentary.}
\label{exp:sports3k}
\vspace{-0.5em}
\end{table}

\subsection{Overall Results}
In Table~\ref{exp:videomme_all} and Table~\ref{exp:sports3k}, we compare the performance of various models on general QA, streaming QA and our LiveSports-3K benchmarks, including advanced proprietary models, SOTA open-source 72B models, and SOTA open-source 7B models.
Despite being initialized from Qwen2-VL-7B-Base, our LiveCC-7B-Instruct outperforms Qwen2-VL-7B-Instruct on the general QA, i.e., VideoMME (64.1 vs. 63.3) and the streaming benchmark, i.e., OVOBench (59.8 vs. 50.4). This demonstrates the strong generalization capabilities of our dataset and training method.
In our proposed LiveSports-3K benchmark, we observe that our three models achieve significant advantages in commentary while maintain competitive on QA, which demonstrates the effectiveness of our method. 

\subsection{Streaming Commentary Capabilities}
\begin{figure}
    \centering
    \includegraphics[width=1.0\linewidth]{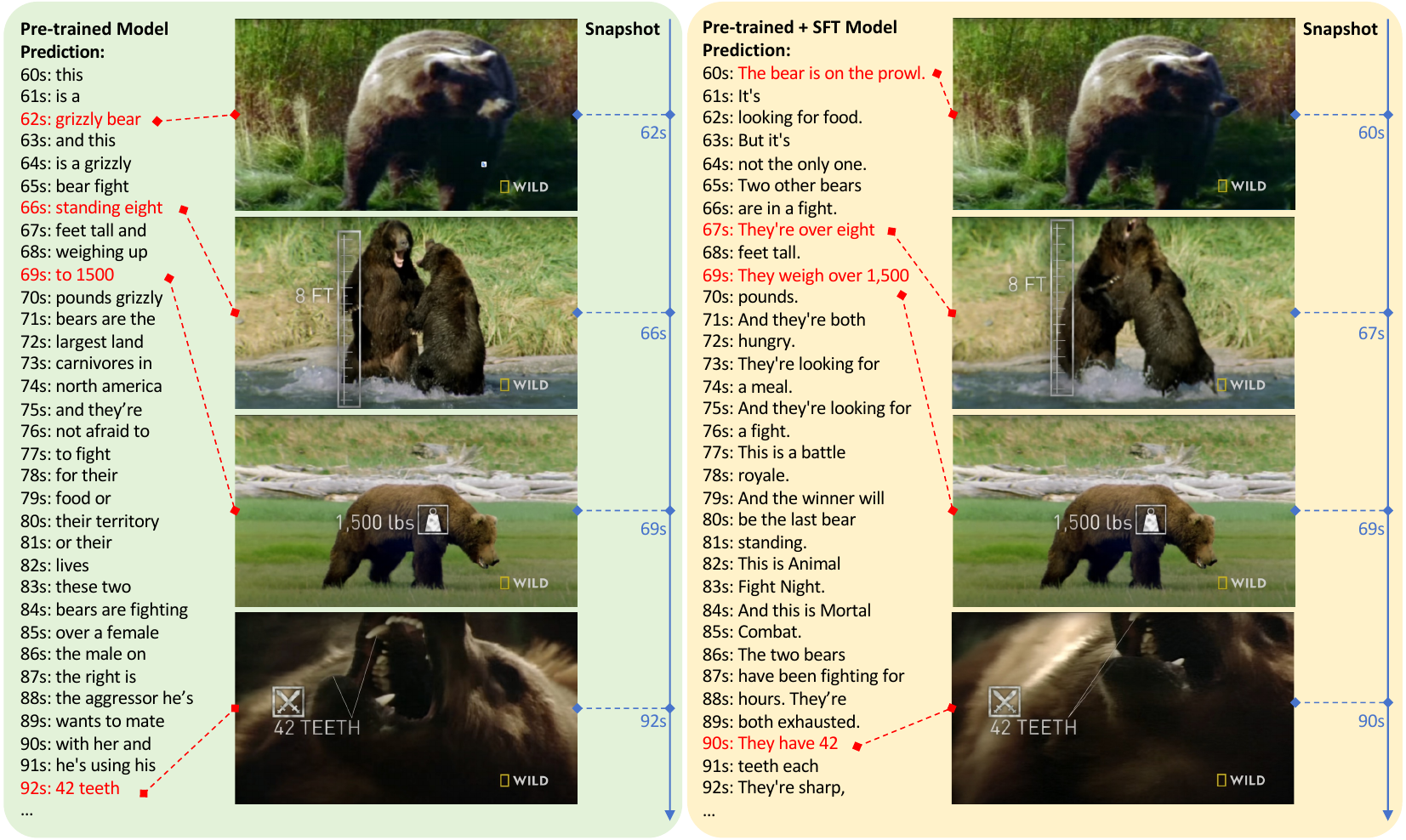}
    \vspace{-1.5em}
    \caption{\textbf{Comparison of pre-trained and instruction tuning enhanced model's predictions on the same video.} This example is sourced from Video-MME~\cite{videomme}, with the YouTube ID \texttt{whksDmTR9YE} featuring animal fights.}
    \label{fig:pt-vs-sft_demo}
    \vspace{-1em}
\end{figure}

Figure~\ref{fig:pt-vs-sft_demo} shows that the pre-trained model can already demonstrate impressive real-time video commentary capabilities.
With SFT, the model further improves formatting (\eg, punctuation, case) and coherence.
More examples are provided in the supplementary material.

%% file: sec/5_conclusion.tex
\section{Conclusion}
In this paper, we investigated the large-scale training of video LLMs using ASR transcripts. We proposed a novel streaming training approach that densely interleaves fine-grained ASR words with their corresponding video frames based on timestamps. Our methodology involved the collection of two datasets: Live-CC-5M for pre-training and Live-WhisperX-526K for instruction tuning. We then developed our streaming pre-training approach, introducing a series of innovative training and inference strategies. Additionally, we designed LiveSports-3K, with two evaluation tracks, LiveSports-3K-CC and LiveSports-3K-QA, which are specifically tailored to assess the model's streaming capabilities. Our extensive experiments demonstrate that our model can perform low-latency commentary for streaming videos and general question answering for holistic video understanding in state-of-the-art performance simultaneously. In future work, we seek methods to train multimodal omni models in streaming.

%% file: sec/X_suppl.tex
\maketitlesupplementary

\begin{figure*}[t]
    \centering
    \includegraphics[width=1.\linewidth]{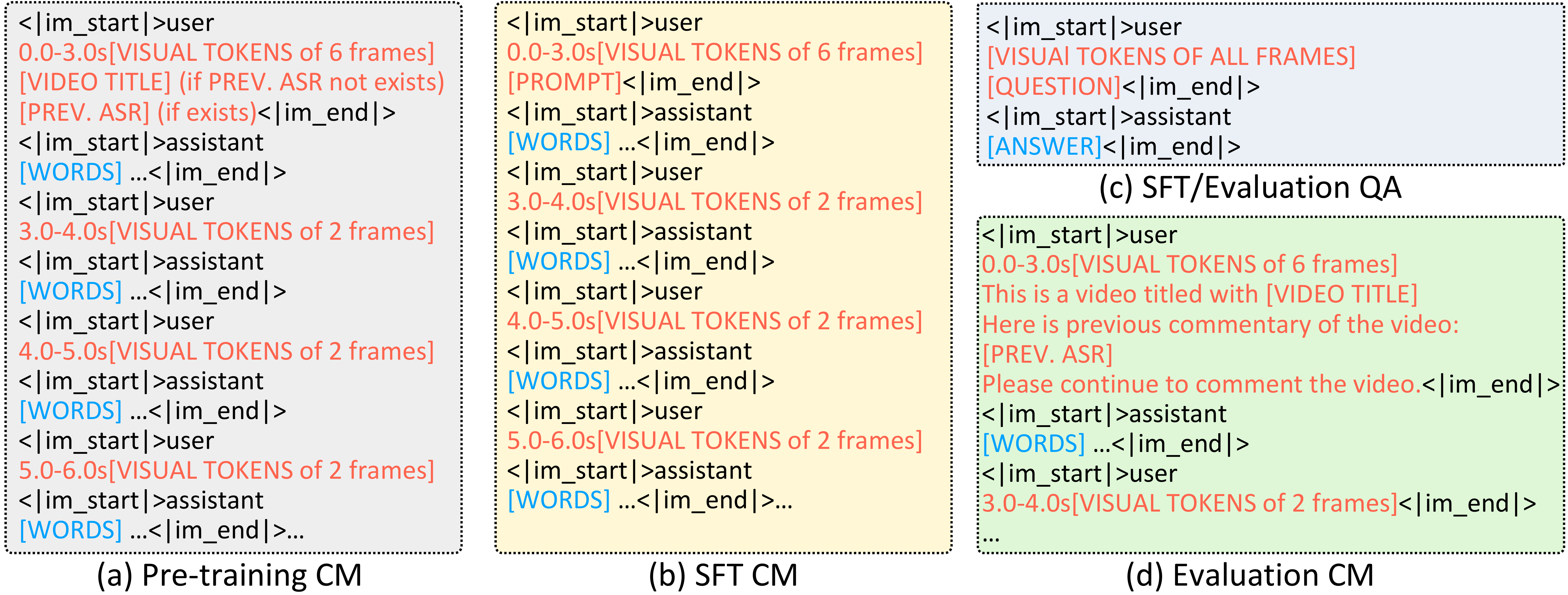}
    \caption{The prompts used during the pre-training instruction-tuning (aka. SFT) stages. CM represents commentary, QA denotes question-answering. For pre-training and instruction-tuning, the previous ASR texts are concatenated to form the context for the live commentary task if they are available. Otherwise, the context is formed by the video title. These contexts are masked during loss calculation. Note that QA data is incorporated exclusively during the instruction-tuning stage. As for inference, we remove the groundtruth in the prompts, \emph{i.e.}, the words followed by a frame or the answer to a multiple-choice question.}
    \label{fig:supp_prompt}
\end{figure*}

\section{Demo}
\begin{figure*}[p]
    \centering
    \begin{subfigure}{\textwidth}
        \centering
        \includegraphics[width=\linewidth]{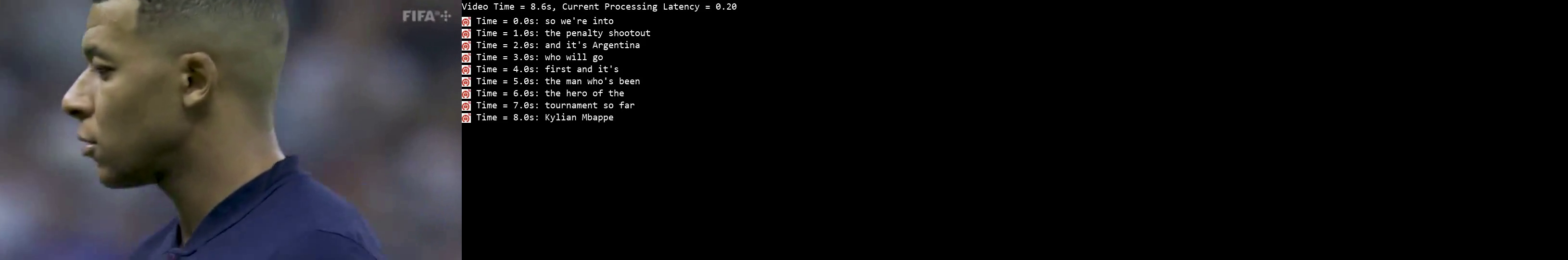}
        \caption{Video Time: 8.6s}
    \end{subfigure}
    \begin{subfigure}{\textwidth}
        \centering
        \includegraphics[width=\linewidth]{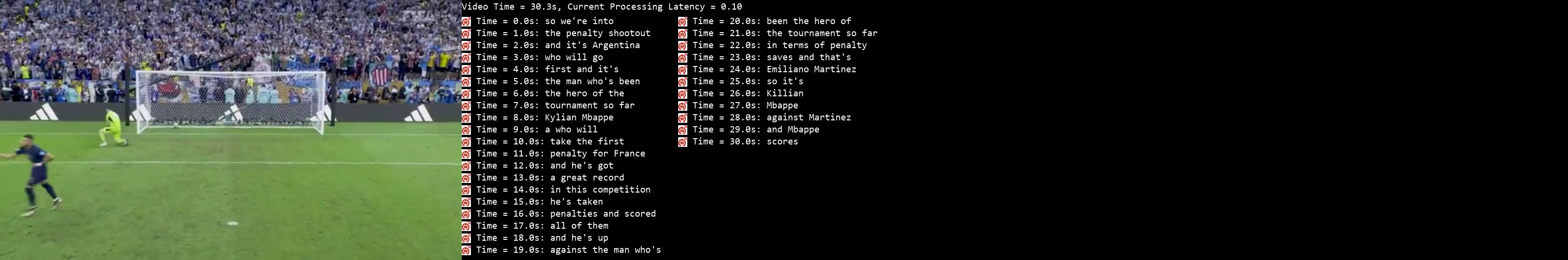}
        \caption{Video Time: 30.3s}
    \end{subfigure}
    \begin{subfigure}{\textwidth}
        \centering
        \includegraphics[width=\linewidth]{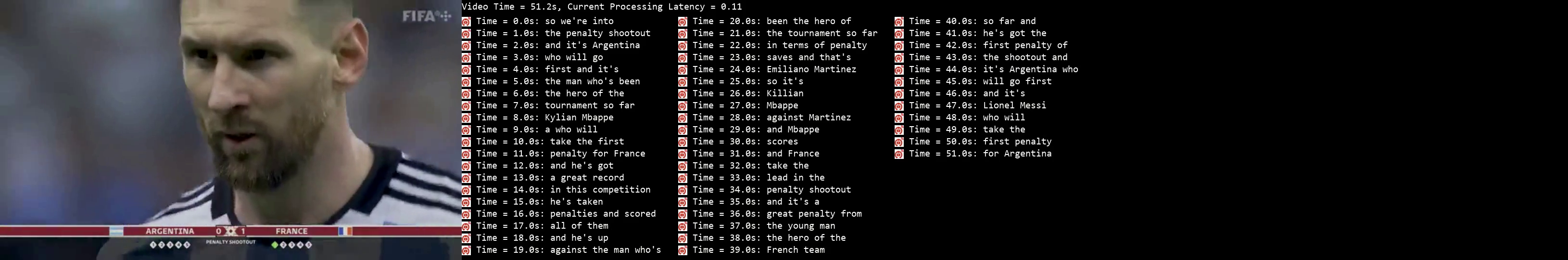}
        \caption{Video Time: 51.2s}
    \end{subfigure}
    \begin{subfigure}{\textwidth}
        \centering
        \includegraphics[width=\linewidth]{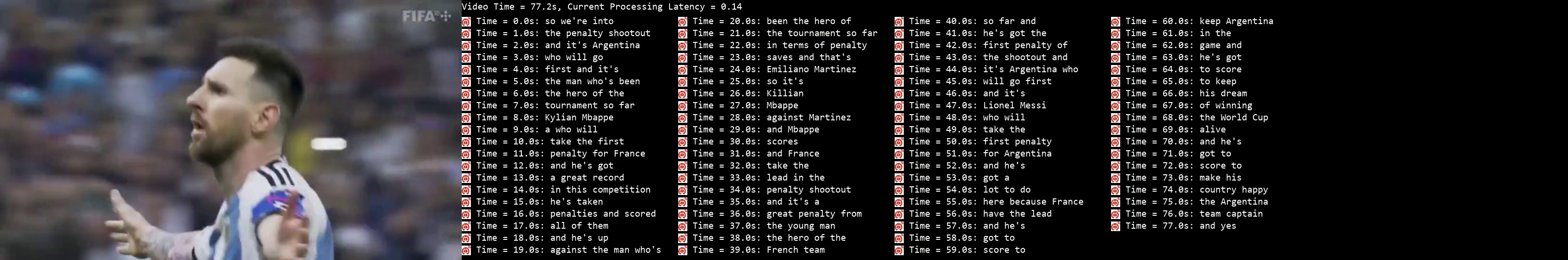}
        \caption{Video Time: 77.2s}
    \end{subfigure}
    \caption{Real-time video commentary demo on unseen YouTube video (\texttt{MCWJNOfJoSM}). The original YouTube title is ``Argentina v France: Full Penalty Shoot-out | 2022 \#FIFAWorldCup Final''. We only give a part of YouTube title ``Full Penalty Shoot-out | 2022 \#FIFAWorldCup Final'' as prompt to avoid information leakage.}
\label{fig:supp_demo1}
\end{figure*}

\begin{figure*}[p]
    \centering
    \begin{subfigure}{\textwidth}
        \centering
        \includegraphics[width=\linewidth]{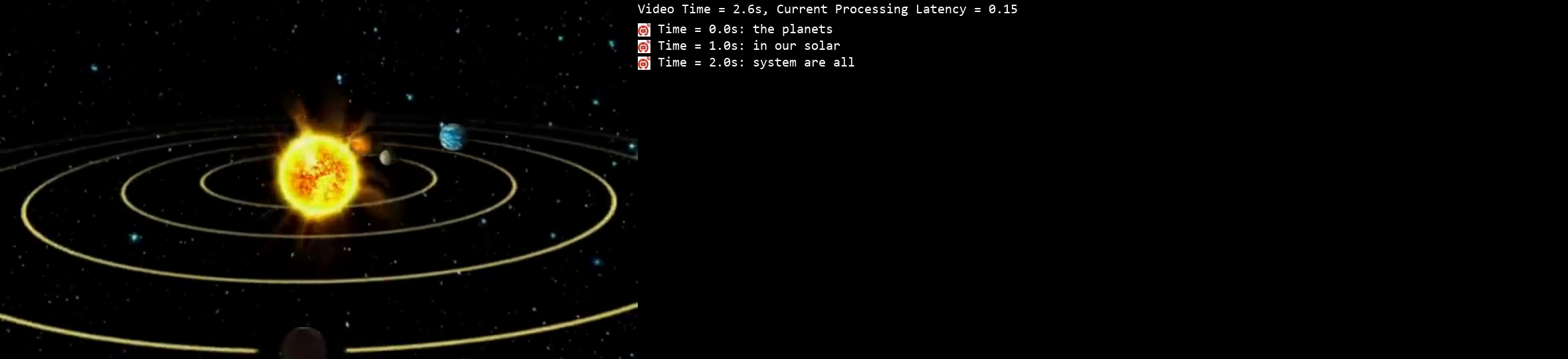}
        \caption{Video Time: 2.6s}
    \end{subfigure}
    \begin{subfigure}{\textwidth}
        \centering
        \includegraphics[width=\linewidth]{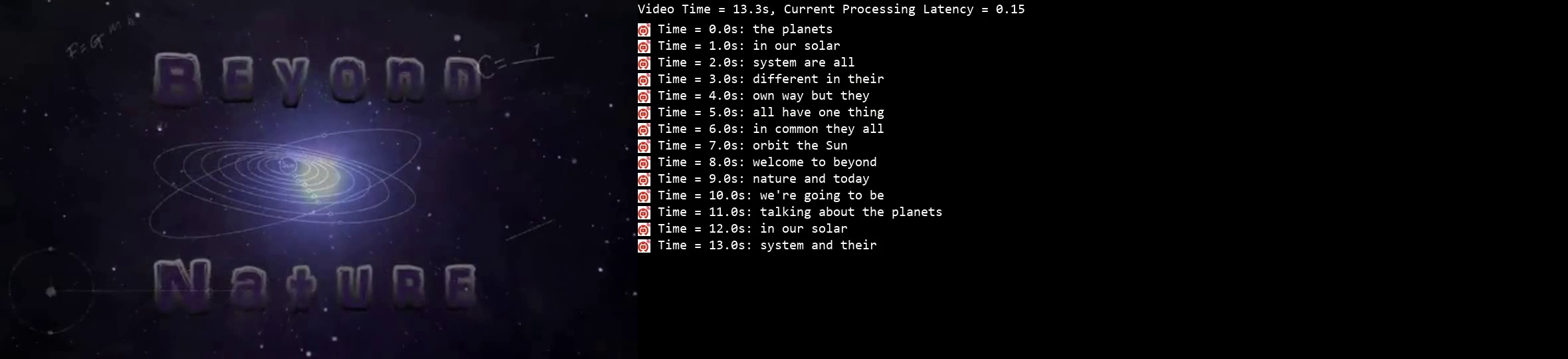}
        \caption{Video Time: 13.3s}
    \end{subfigure}
    \begin{subfigure}{\textwidth}
        \centering
        \includegraphics[width=\linewidth]{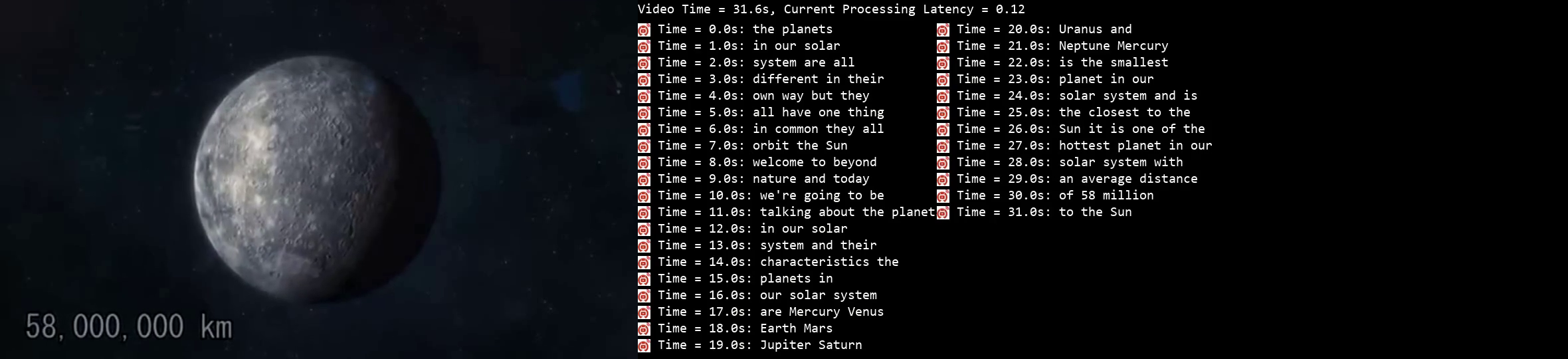}
        \caption{Video Time: 31.6s}
    \end{subfigure}
    \begin{subfigure}{\textwidth}
        \centering
        \includegraphics[width=\linewidth]{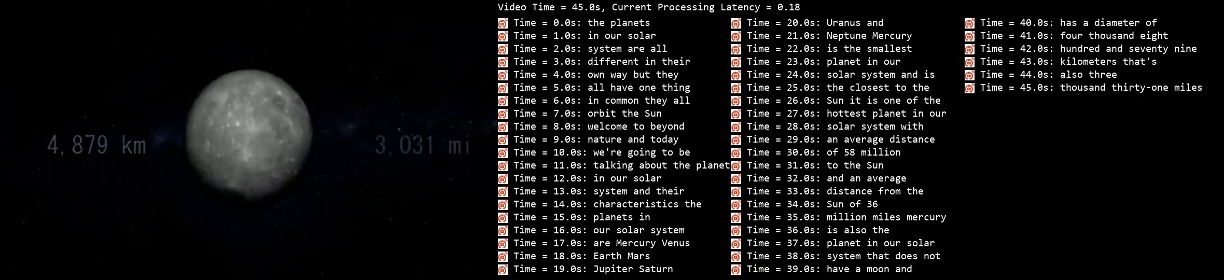}
        \caption{Video Time: 45.0s}
    \end{subfigure}
    \caption{Real-time video commentary demo on unseen YouTube video (\texttt{lcZTcfdZ3Ow}). We give the YouTube title ``The Planets In Our Solar System'' as prompt.}
\label{fig:supp_demo2}
\end{figure*}

\begin{figure*}[p]
    \centering
    \begin{subfigure}{\textwidth}
        \centering
        \includegraphics[width=\linewidth]{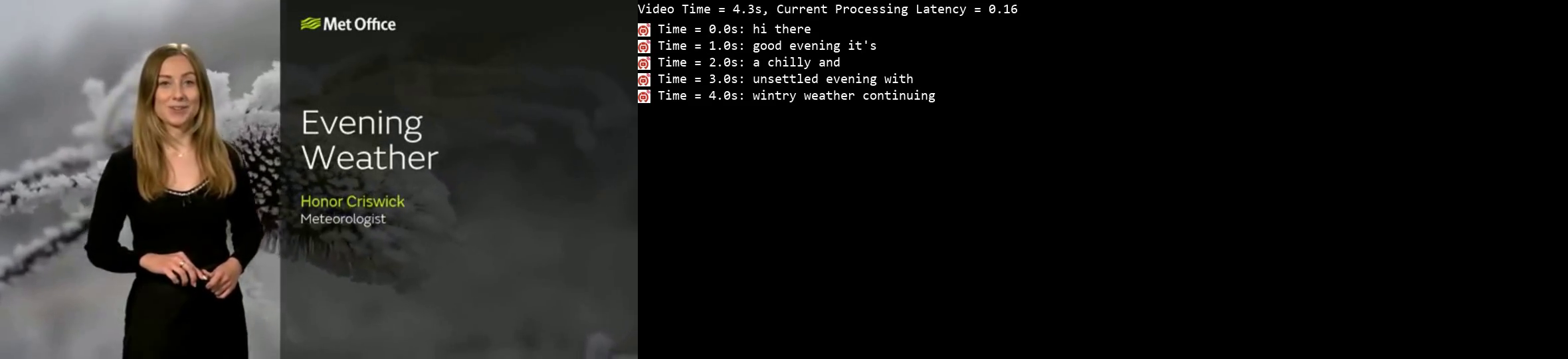}
        \caption{Video Time: 4.3s}
    \end{subfigure}
    \begin{subfigure}{\textwidth}
        \centering
        \includegraphics[width=\linewidth]{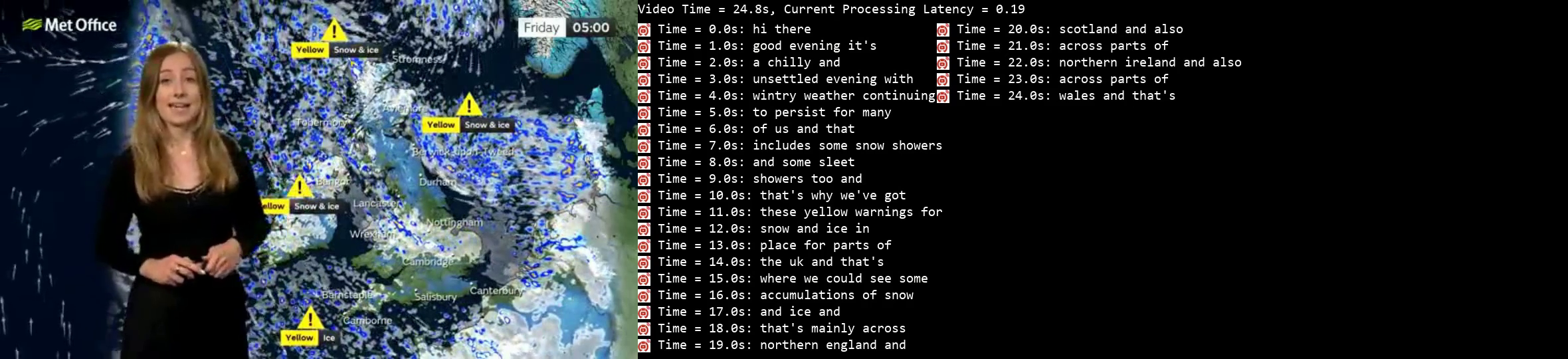}
        \caption{Video Time: 24.8s}
    \end{subfigure}
    \begin{subfigure}{\textwidth}
        \centering
        \includegraphics[width=\linewidth]{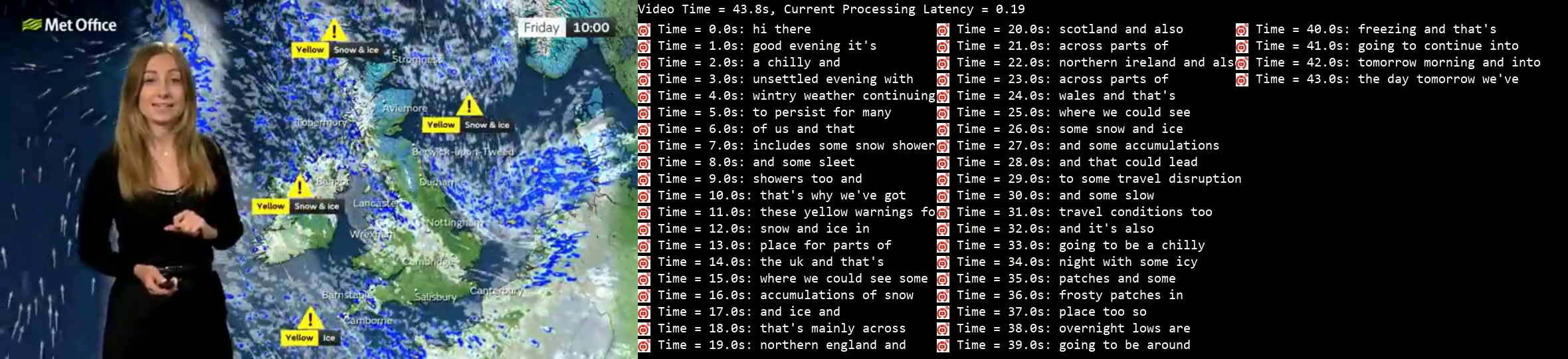}
        \caption{Video Time: 43.8s}
    \end{subfigure}
    \begin{subfigure}{\textwidth}
        \centering
        \includegraphics[width=\linewidth]{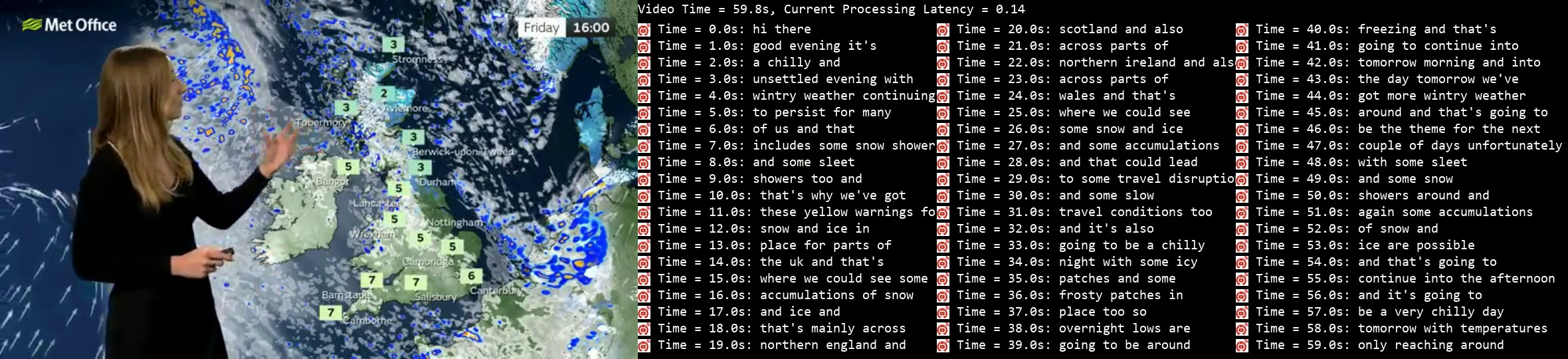}
        \caption{Video Time: 59.8s}
    \end{subfigure}
   \caption{Real-time video commentary demo on unseen YouTube video (\texttt{8XajZdrCDsk}). The original YouTube title is ``21/11/24 - Wintry weather perservering - Evening Weather Forecast UK – Met Office Weather''. We only give ``21/11/24 - Wintry weather perservering'' as prompt to avoid information leakage.}
\label{fig:supp_demo3}
\end{figure*}

\begin{figure*}[p]
    \centering
    \begin{subfigure}{\textwidth}
        \centering
        \includegraphics[width=\linewidth]{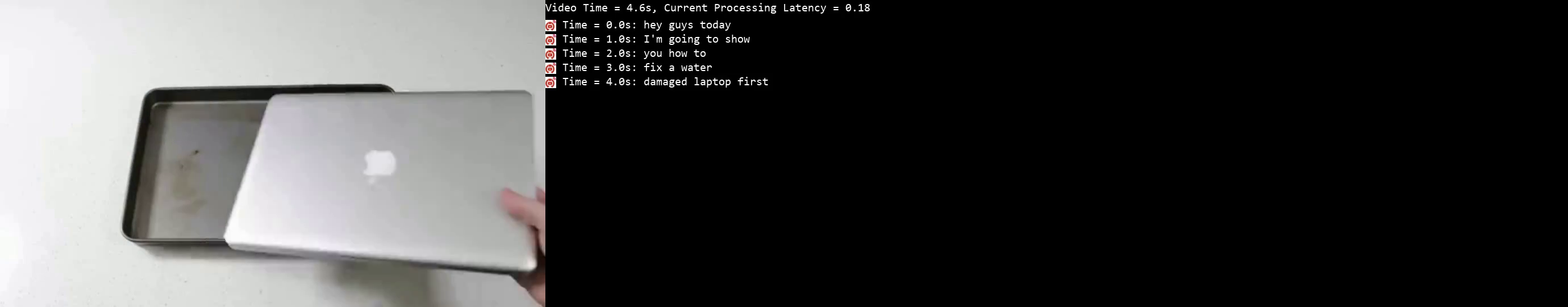}
        \caption{Video Time: 4.6s}
    \end{subfigure}
    \begin{subfigure}{\textwidth}
        \centering
        \includegraphics[width=\linewidth]{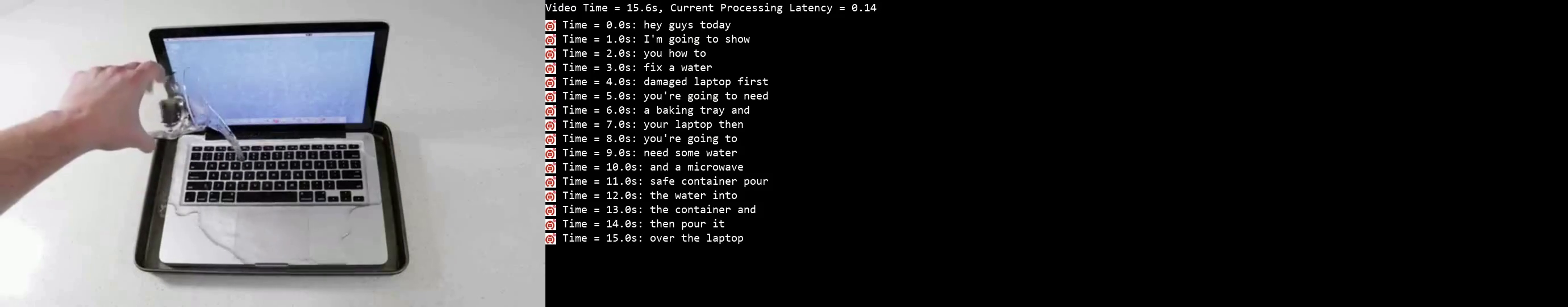}
        \caption{Video Time: 15.6s}
    \end{subfigure}
    \begin{subfigure}{\textwidth}
        \centering
        \includegraphics[width=\linewidth]{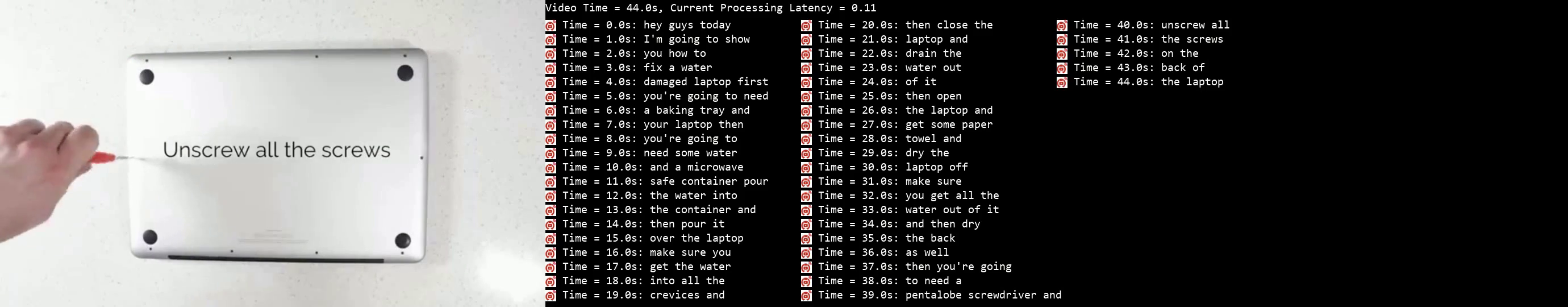}
        \caption{Video Time: 44.0s}
    \end{subfigure}
    \begin{subfigure}{\textwidth}
        \centering
        \includegraphics[width=\linewidth]{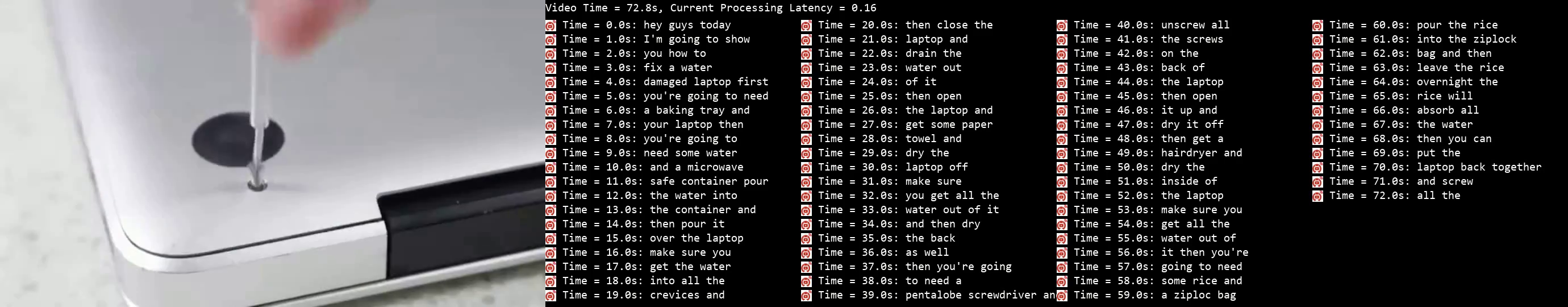}
        \caption{Video Time: 72.8s}
    \end{subfigure}
   \caption{Real-time video commentary demo on unseen YouTube video (\texttt{115amzVdV44}). We give the YouTube title ``How To Fix a Water Damaged Laptop'' as the prompt.}
\label{fig:supp_demo4}
\end{figure*}

This section showcases four demo videos to demonstrate the capability of our LiveCC-7B-Instruct to provide real-time commentary in real-world videos across different domains, including sports (football), science (astronomy), news (weather forecast), and instructional (computer repair) videos. As illustrated in Figure~\ref{fig:supp_demo1}-\ref{fig:supp_demo4}, in the first demo, our LiveCC-7B-Instruct model correctly recognizes all exact penalty timings, highlighting its strong temporal perception abilities. By leveraging the extensive world knowledge gained from watching millions of YouTube videos, our model accurately reports the name of the related player. The second demo showcases the model’s ability to comment beyond sports by precisely presenting astronomy knowledge and demonstrating good OCR capability to read large numbers. The third demo further reveals its fine-grained temporal understanding capability, as evidenced by its real-time commentary on subtle changes in weather maps. The final demo demonstrates that our model is also capable of generating a tutorial to guide users, revealing its potential to serve as a real-time assistant. 

\section{Implemetation Details}

\subsection{Prompt Template}

In this section, we detail the prompt designs used during the pre-training, instruction tuning, and inference stages. As shown in Figure~\ref{fig:supp_prompt}(a) and (b), the video title and previously transcribed ASR text are provided as contextual information during pre-training but are omitted during SFT. For the first round of vision token extraction, we use a 3-second video clip, followed by 1-second clips in subsequent rounds. Given a frame rate of 2 FPS, this corresponds to 6 and 2 frames, respectively. For QA-style data used exclusively in SFT, illustrated in Figure~\ref{fig:supp_prompt}(c), we adopt the input format of Qwen2-VL-Instruct~\cite{qwen2vl}, which is also used during evaluation. However, for real-time commentary evaluation, we follow the format in Figure~\ref{fig:supp_prompt}(d), where the video title and previous ASR transcripts are included to ensure consistency with other non-streaming baselines.

\subsection{Win Rate Computation on LiveSports-3K}
In this section, we present the detailed process for computing the win rate on LiveSports-3K-CC. To start, we categorize the models into two groups based on their inference schemes:
\textbf{i}) \textbf{Clip-wise caption models}, including GPT-4o~\cite{gpt4o}, Gemini-1.5-Pro~\cite{gemini}, LLaVA-OV-7/72B~\cite{llavaov}, LLaVA-Video-7/72B~\cite{llavavideo}, Qwen2-VL-7/72B-Instruct~\cite{qwen2vl}, Qwen2.5-VL-7/72B-Instruct~\cite{qwen2.5vl} and Qwen2.5-Omni-7B~\cite{qwen2.5omni}.
\textbf{ii}) \textbf{Frame-wise streaming model}, \emph{i.e.}, our  LiveCC-7B-Base, Qwen2-VL-7B-LiveCCInstruct, LiveCC-7B-Instruct.

For clip-wise caption models, we directly input the overall event clips, perform a \textbf{one-time} generation, and use the generated response as the commentary. To ensure stylistic consistency and fair evaluation, the same prompt context as that shown in Figure~\ref{fig:supp_prompt}(d) is applied across all models. We use the video commentary from GPT-4o-08-06~\cite{gpt4o}  serves as the baseline for comparison with other models.
For our models, we adopt \textbf{streaming} inference shown in Figure~\ref{fig:supp_prompt}(d), where commentary is generated frame by frame. The generated tokens are then concatenated to form the complete commentary, which is evaluated for quality.

For evaluation, we also prompt GPT-4o-08-06~\cite{gpt4o} to assess whether a given commentary surpasses that of GPT-4o-08-06. The evaluation is based on two key criteria:
(\textbf{i}) Semantic Alignment, \emph{i.e.}, consider which text conveys the same meaning, details, and key points as the groundtruth ASR transcript, with minimal deviation.
(\textbf{ii}) Stylistic Consistency, \emph{i.e.}, assesses which text maintains a tone, word choice, and structure similar to the ground-truth transcript.
The overall prompt is written as:
\definecolor{bgcolor}{rgb}{0.95,0.95,0.95}

\lstset{
  backgroundcolor=\color{bgcolor},
  basicstyle=\footnotesize\ttfamily,
  breaklines=true,
  breakindent=0pt,
  showstringspaces=false,
  frame=single
}
\begin{lstlisting}[language=, numbers=none]
You are an expert in video commentary.  Your task is to review two commentaries (Commentary A and Commentary B), and select the one that better aligns with the human commentary. You should consider the criteria:
1. Semantic Alignment: The commentary should convey the same meaning, details, and key points as the human commentary.
If the above criteria is not enough to judge, then consider:
2. Stylistic Consistency: The commentary should maintain a tone, word choice, and structure similar to the human commentary.
---Commentary A---
{a_pred}
----------
---Commentary B---
{b_pred}
----------
---Human Commentary---
{gt_asr}
----------
Your response should be "Commentary A is better aligned with the human commentary" or "Commentary B is better aligned with the human commentary".
\end{lstlisting}
The final win rate is calculated as the proportion of instances where GPT-4o~\cite{gpt4o} selects the model’s response over the baseline. To mitigate positional bias in GPT's responses, each prompt is evaluated \textit{twice} with the positions of the tested model and baseline text swapped.

\section{Additional Experiments}
\subsection{Response Latency}

\begin{table}[t]
\centering
\resizebox{1.\columnwidth}{!}{
\begin{tabular}{lccc}
\toprule
Model           & Latency & Input & Inf. Type \\
\midrule
LLaVA-Video-72B~\cite{llavavideo} & 20.51s                 & Clip  & Captioning        \\
LLaVA-Video-7B~\cite{llavavideo}  & 5.62s             & Clip  & Captioning        \\
\baseline{LiveCC-7B-Instruct}  & \baseline{\textbf{0.17s}}               & \baseline{Frame} & \baseline{Streaming}     \\
\bottomrule
\end{tabular}}
\caption{The response latency comparison between LLaVA-Video-7/72B and our LiveCC-7B. Inf. is short for Inference.}
\label{tab:supp_latency}
\end{table}

\begin{figure}
    \centering
    \includegraphics[width=1.\linewidth]{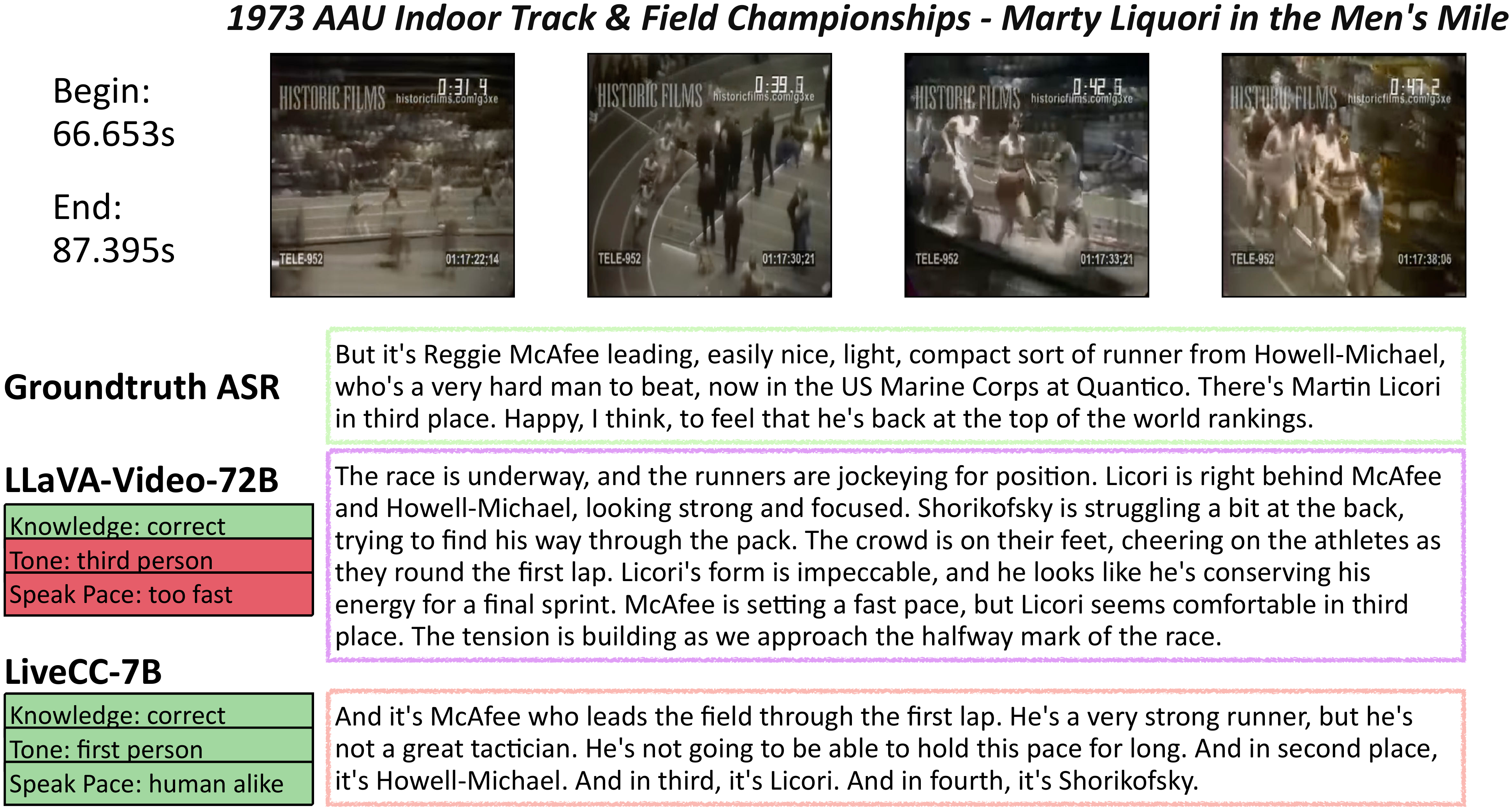}
    \caption{The comparison between the commentary generated by LLaVA-Video-72B and our LiveCC-7B-Instruct.}
    \label{fig:supp_vis}
\end{figure}
To highlight the efficiency of our streaming model, we present the response latency of LLaVA-Video-7B/72B alongside our model in Table~\ref{tab:supp_latency}. Response latency is defined as the time a user waits to see the model’s output, a critical factor affecting user experience. Since the LLaVA-Video series are trained in a captioning style, requiring a full clip as input rather than a single frame, their response latency is significantly higher than that of our model. Notably, LiveCC not only achieves lower latency but also delivers high-quality commentary (see Table~\ref{exp:sports3k}).

\subsection{Commentary Quality}

We analyzed the quality of the generated content, as shown in Figure~\ref{fig:supp_vis}. Benefiting from training on millions of ASR-transcribed videos, our model produces commentary that is more aligned with human preferences in terms of tone and speaking pace, while maintaining accurate event understanding. In contrast, the LLaVA-Video-72B, although capable of correctly describing the event, falls short in emulating human-like commentary. 


